\documentclass[acmtog, nonacm]{acmart} %

\AtBeginDocument{%
  }

\usepackage{booktabs,graphicx}
\usepackage[table]{xcolor}
\usepackage{algorithm}
\usepackage[noend]{algpseudocode}

\usepackage[capitalise]{cleveref}

\definecolor{Teal}{HTML}{09C8AE}

\begin{document}

\title{Differentiable Voronoi Ray Tracing Beyond Rasterization Speeds}

\author{Bernardo Taveira}
\orcid{0009-0006-4592-2289}
\affiliation{
  \institution{Chalmers University of Technology}
  \country{Sweden}
  \city{Gothenburg}
}
\affiliation{%
 \institution{Zenseact}
 \country{Sweden}
 \city{Gothenburg}
}
\email{bernardo.taveira@chalmers.se}

\author{Carl Lindström}
\orcid{0009-0006-3563-8946}
\affiliation{
  \institution{Chalmers University of Technology}
  \country{Sweden}
  \city{Gothenburg}
}
\affiliation{%
 \institution{Zenseact}
 \country{Sweden}
 \city{Gothenburg}
}
\email{carl.lindstrom@chalmers.se}
\author{Joakim Johnander}
\orcid{0000-0003-2553-3367}
\affiliation{%
 \institution{Zenseact}
 \country{Sweden}
 \city{Gothenburg}
}
\affiliation{
  \institution{Linköping University}
  \country{Sweden}
  \city{Linköping}
}
\email{joakim.johnander@zenseact.com}
\author{Fredrik Kahl}
\orcid{0000-0001-9835-3020}
\affiliation{
  \institution{Chalmers University of Technology}
  \country{Sweden}
  \city{Gothenburg}
}
\email{fredrik.kahl@chalmers.se}

\renewcommand{\shortauthors}{B. Taveira et al.}

\def\methodname{VoroTracing}

\begin{teaserfigure}
    \centering
    \includegraphics[width=\textwidth]{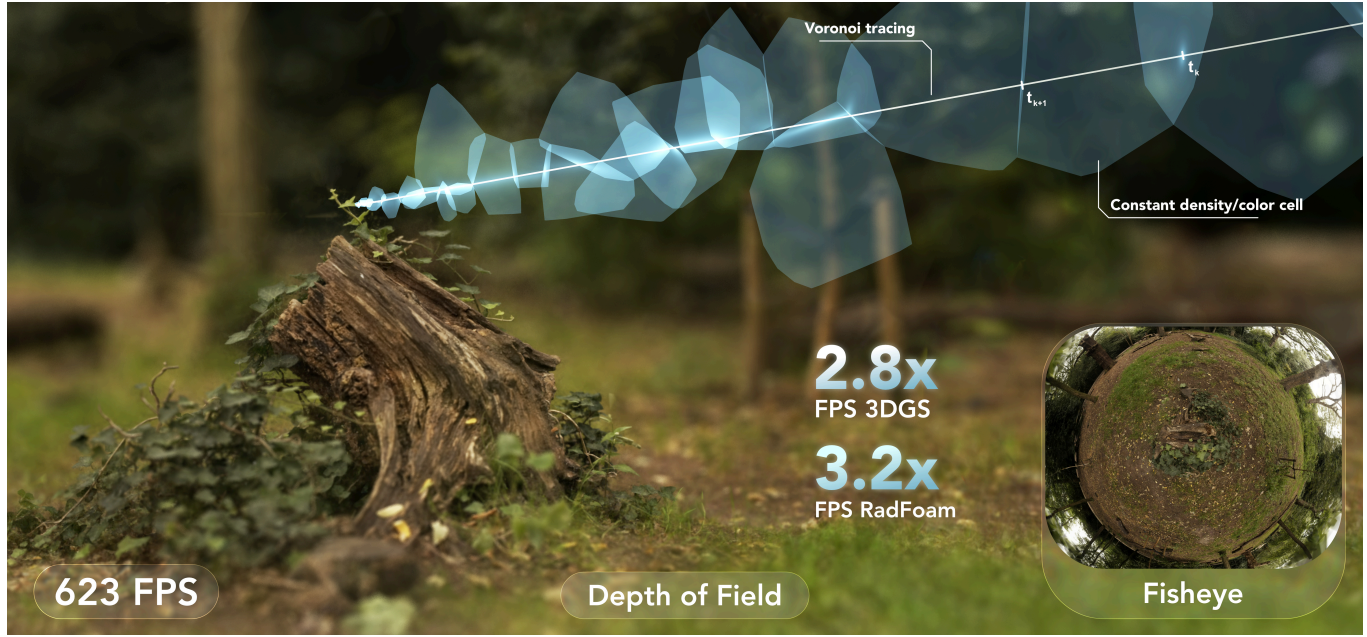}
    \caption{\textbf{Ray-based rendering with \methodname{}.} The main image is a shallow-depth-of-field rendering of the Garden scene. The highlighted path visualizes an actual camera ray traversing the scene's Voronoi diagram from cell to cell. The light blue regions show the projected footprints of the 3D cells traversed by the ray. The lower-right inset shows a fisheye rendering of the same scene from an overhead viewpoint. Both effects are produced by the same trained representation through changes to ray generation and sampling.}
    \label{fig:cover}
    \Description{Teaser showing a shallow-depth-of-field rendering of a garden scene. A highlighted camera ray passes through a sequence of Voronoi cells, whose projected footprints are shown as light blue regions. A lower-right inset shows an overhead fisheye view of the same scene.}
\end{teaserfigure}
\begin{abstract}
  Real-time novel view synthesis is dominated by rasterized explicit primitives. These projection-based pipelines provide high throughput but require specialized extensions for non-pinhole effects such as distortion, rolling shutter, and depth of field. Ray-based rendering expresses these effects naturally but is generally assumed too slow for competitive real-time rendering. We analyze the factors governing throughput in differentiable Voronoi ray tracing and identify traversal length, per-cell work, and memory locality as principal determinants. Guided by this, we introduce \methodname{}, which co-designs the scene representation, optimization, and GPU execution to reduce these costs. Compact octahedral appearance textures reduce memory traffic, while surface-concentrated opacity promotes early termination. The fixed-budget representation is optimized without pruning or densification and rendered with a GPU implementation designed for coherent traversal. On Mip-NeRF~360, \methodname{} renders at 623 FPS on an RTX~5090, providing $3.2\times$ the throughput of the fastest prior ray-based method and $2.8\times$ that of 3D Gaussian Splatting, while maintaining competitive reconstruction quality. Our renderer supports fisheye, rolling-shutter, motion-blur, and depth-of-field effects through ray generation and sampling, requiring no specialized rasterization. These results show that real-time throughput can be achieved with the flexibility of ray-based rendering. We release our source code, see {\color{Teal}\url{https://research.zenseact.com/publications/vorotracing}}
\end{abstract}



\maketitle

\section{Introduction}
\label{sec:introduction}

Rasterization is one of the defining approximations of real-time computer graphics. By replacing general light transport with projection, visibility, and local shading, it enabled interactive 3D graphics long before physically complete rendering was practical. Its success came not from being the most general image formation model, but from an exceptional co-design of representation, algorithms, and graphics hardware. Decades of graphics systems have been organized around this projection-based pipeline.

Novel view synthesis saw a similar evolution. The task is not to render a known scene, but to reconstruct a scene representation from images and render it from new viewpoints. Neural radiance fields~\cite{mildenhall2021nerf} approached this problem through differentiable ray-based volume rendering. This made image formation flexible and mathematically direct, but at a high computational cost, since each camera ray required many samples and repeated neural evaluations. 3D Gaussian Splatting (3DGS)~\cite{kerbl20233d} changed the practical trajectory of the field by bringing rasterization techniques into novel view synthesis. Through projected primitives, tile sorting and alpha compositing, 3DGS introduced interactive rendering speeds to the novel view synthesis field. The speedup was large enough that much of real-time novel view synthesis has since been organized around rasterized splats and related projected primitives.

Following this paradigm shift, recent reconstruction methods have increasingly extended rasterized splatting beyond the standard pinhole-camera setting. Fisheye and distorted cameras typically require modified projection~\cite{liao2024fisheye, deng2025self, ren2026unigaussian, wu20253dgut}. Rolling-shutter cameras require time aware projection or, more generally, a time varying image formation model~\cite{wu20253dgut, hess2025splatad, seiskari2024gaussian}. Depth of field requires extra blurring steps~\cite{wang2024cinematic, wang2024dof, shen2025dof}. These extensions substantially broaden the applicability of splatting-based reconstruction, but they also increase complexity of the original rasterization pipeline, introduce further approximations, and typically come at the cost of rendering speed. Recent ray-traced and hybrid Gaussian methods make this trade-off explicit, using ray tracing to support general image formation while retaining rasterization for speed~\cite{moenne20243d, govindarajan2026power, mai2026radiance}.

This raises the question of whether real-time novel view synthesis must be built around projection-based rendering, or around representations that preserve a rasterized path for speed alongside ray tracing. The common view is that rays provide flexibility and rasterization provides speed. We argue that, for the explicit radiance-field setting studied here, this trade-off is not inherent. In this work, we show that a ray-based renderer designed around traversal cost, memory traffic, opacity, and GPU execution can retain the flexibility of rays while matching or exceeding the throughput of strong rasterized baselines.

Radiant Foam~\cite{govindarajan2025radiant} provides an important starting point. It represents the scene explicitly as a Voronoi partition whose cells store density and appearance, and replaces repeated BVH intersections with traversal through this partition. Once a ray is inside a cell, the next cell is found through local adjacency, making the cost depend primarily on the number of cells visited rather than on the total number of primitives. This removes a major obstacle to fast ray tracing. The remaining cost is dominated by what happens at each ray-cell interaction, including the amount of appearance data loaded, the number of semi-transparent cells composited before termination, and the optimization procedure needed to obtain a useful cell distribution.

We revisit the full Voronoi ray-tracing pipeline from this perspective. Our method reduces the cost of each ray-cell interaction with a compact texture appearance model, encourages opacity to concentrate on surfaces so rays terminate quickly, and uses a scale-invariant density parameterization that avoids biases caused by varying cell sizes. Together, these changes make the cost of rendering depend on a short sequence of lightweight cell evaluations rather than on repeated global intersection queries or expensive per-cell appearance loads. The same design philosophy extends to training. Rather than relying on pruning, densification, opacity resets, progressive downsampling, or other structural heuristics, we initialize a fixed dense set of sites from dense image matching and optimize the representation directly. This does not imply that adaptive cell insertion is unimportant. Better densification strategies may further improve future systems. Our point is that they should not be necessary to obtain high-quality reconstructions from a well-posed representation.

\begin{figure}[t]
  \centering
  \includegraphics[width=0.95\columnwidth]{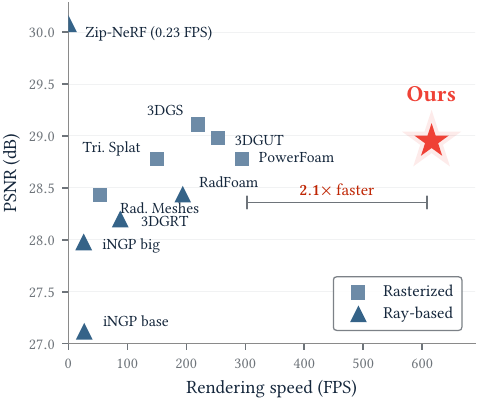}
  \caption{\textbf{Speed--quality trade-off on Mip-NeRF~360.} Rendering speed
    (RTX~5090) versus reconstruction quality, averaged over the seven scenes.
    \methodname{} renders $2.1\times$ faster than the fastest prior baseline while
    matching the best rasterized quality to within $0.2$\,dB. Marker shape indicates the
    renderer used for the reported timing.}
  \label{fig:pareto}
  \Description{Scatter plot of average PSNR against rendering speed for ray-based and rasterized novel-view-synthesis methods, with our method in the high-speed, high-quality region.}
\end{figure}

On Mip-NeRF~360, our renderer reaches $623$ FPS on an RTX~5090 (\cref{fig:pareto}). This is $3.2\times$ faster than the fastest prior ray-based method in our comparison, while improving over Radiant Foam across PSNR, SSIM, and LPIPS. Against rasterized methods, it is the only ray-based method in our comparison and renders roughly $2\times$ faster than all listed rasterizers, while remaining competitive in appearance quality. These results show that, in this novel-view-synthesis setting, returning to rays does not require giving up real-time performance. In summary, our main contributions are:
\begin{itemize}
    \item We show that differentiable ray tracing can match or exceed the throughput of rasterized novel-view-synthesis baselines, by designing around traversal cost, memory traffic, and early termination.
    \item We replace per-cell spherical harmonics with octahedral surface and view-dependent textures, reducing the number of appearance values loaded at each ray-cell interaction while allowing spatial detail within a cell.
    \item We introduce a surface-concentrated opacity formulation, together with a scale-invariant density parameterization, encouraging transparent free space and early termination at compact surfaces.
    \item We optimize inference with spatial cell ordering, warp-coherent ray scheduling, aligned texture loads, and low-contribution cell skipping.
    \item We use a simple fixed-budget training setup initialized from dense correspondences, avoiding pruning, densification, progressive downsampling, and multi-stage schedules.
\end{itemize}

\section{Related Work}

\paragraph{Real-time novel view synthesis.}
Neural radiance fields~\cite{mildenhall2021nerf} established differentiable volume rendering along camera rays as a flexible formulation for novel view synthesis, but rendering remained expensive, with many samples and neural evaluations required per pixel. Subsequent work reduced this cost through explicit or hybrid representations, including multiresolution hash encodings~\cite{muller2022instant}, baked radiance grids~\cite{hedman2021baking}, and explicit radiance fields based on octrees or sparse voxels~\cite{yu2021plenoctrees,fridovich2022plenoxels}. 3D Gaussian Splatting~\cite{kerbl20233d} then shifted the practical focus of real-time novel view synthesis toward rasterized explicit primitives, combining anisotropic Gaussians, adaptive density control, tile sorting, and spherical-harmonic appearance into an interactive, high-quality renderer. Recent methods extend this rasterized direction to surface-aligned, triangular, or bounded volumetric primitives~\cite{huang20242d,held2026triangle,von2026linprim}. Our work keeps an explicit primitive representation but revisits whether the renderer itself must be projection based to be fast.

\paragraph{Ray-based and hybrid rendering.}
A recurring limitation of rasterized splatting is its reliance on projection. Departures from the ideal pinhole camera often require specialized projection or image-formation modules for fisheye and distorted lenses~\cite{liao2024fisheye,deng2025self,ren2026unigaussian,wu20253dgut}, rolling shutter and motion blur~\cite{seiskari2024gaussian,hess2025splatad}, and depth of field~\cite{wang2024cinematic,wang2024dof,shen2025dof}. Ray-based image formation instead expresses these effects through ray generation. 3D Gaussian Ray Tracing~\cite{moenne20243d} traces Gaussian particle scenes with hardware-accelerated ray tracing, supporting distorted cameras and secondary rays while still relying on BVH traversal. Radiant Foam~\cite{govindarajan2025radiant} replaces repeated BVH queries with adjacency traversal through a Voronoi partition, approaching the speed of 3DGS while remaining fully ray-based. \citet{taveira2026paragram} scale this construction to larger cell counts, improving fidelity with diminishing returns, while Power Foam~\cite{govindarajan2026power} and Radiance Meshes~\cite{mai2026radiance} further blur the boundary between rasterization and ray tracing through power diagrams and tetrahedral meshes. We build on Voronoi ray tracing, targeting the per-cell costs that remain after global intersection queries have been removed.

\paragraph{Appearance inside explicit primitives.}
The appearance stored in each primitive is where representation quality and rendering cost meet. 3DGS~\cite{kerbl20233d} and Radiant Foam~\cite{govindarajan2025radiant} store view-dependent color using spherical harmonics. This representation is compact but limited to smoothly varying, low-frequency directional effects, and its coefficient count grows quadratically with angular frequency. Ref-NeRF~\cite{verbin2024ref} shows that structuring the view-dependent term around the reflected direction improves specular appearance, and Spec-Gaussian~\cite{yang2024specgaussian} reaches a similar conclusion for splatting by replacing spherical harmonics with anisotropic spherical Gaussians. These works primarily address angular variation. A complementary issue is spatial variation within a primitive. With a single directional color field, texture detail across a footprint must be represented by adding more primitives. Textured Gaussians~\cite{chao2025textured} relax this by mapping a small texture onto each Gaussian. In Voronoi tracing this trade-off is especially important because appearance lookup is part of the per-cell traversal loop. We use octahedral maps~\cite{cigolle2014survey} to add spatial and directional detail while keeping appearance traffic bounded.

\paragraph{Initialization and adaptive density control.}
Explicit radiance-field methods depend heavily on where primitives are placed and how they are allowed to change during optimization. 3DGS~\cite{kerbl20233d} relies on adaptive density control to clone, split, and prune primitives, and Radiant Foam~\cite{govindarajan2025radiant} similarly couples pruning, densification, and progressive image resolution to reach a usable cell distribution. These heuristics are effective but scene-dependent and difficult to tune. EDGS~\cite{kotovenko2026edgs} shows that dense, geometry-aware initialization can largely remove the need for repeated densification, and dense correspondence methods such as RoMa v2~\cite{edstedt2025roma} provide a practical source of this initial structure. We use dense correspondences to start from a fixed cell budget, so changes in quality or speed can be attributed to the representation and renderer rather than to cells being inserted or removed during training.

\paragraph{Surface concentration.}
A separate line of work reduces the volumetric ambiguity of splatting by pushing the representation toward opaque surfaces. Surface-aligned and planar primitives such as 2D Gaussian Splatting~\cite{huang20242d} and triangle splatting~\cite{held2026triangle} constrain appearance to thin geometry, and recent work drives these primitives to be fully opaque, optimizing opaque triangles or meshes that map directly onto standard graphics pipelines~\cite{held2025trianglesplattingplus,held2025meshsplatting}. Surface-aligned Gaussian methods similarly add regularizers that pull the representation onto a well-defined surface~\cite{guedon2024sugar,yu2024gaussian}. For ray traversal, surface concentration also has a direct computational consequence, since compact opacity reduces the number of semi-transparent regions traversed and composited before a ray terminates. The distortion regularizer introduced in Mip-NeRF~360~\cite{barron2022mip} penalizes opacity spread along rays, providing a useful way to encourage compact support without prescribing a depth target.

\section{Preliminaries}
\label{sec:preliminaries}

We first review the Voronoi tracing framework introduced by Radiant Foam~\cite{govindarajan2025radiant}, focusing on the components that our method modifies or extends. We also establish the notation used throughout the paper.

\subsection{Voronoi Ray Traversal}

Radiant Foam~\cite{govindarajan2025radiant} replaces repeated BVH intersection queries with traversal through a Voronoi diagram. The diagram partitions space into cells of constant density and appearance, so a ray can be advanced by walking from cell to cell through local adjacency rather than repeatedly querying a global acceleration structure.

\paragraph{Voronoi Diagram.}
The Voronoi diagram and its more popular dual, the Delaunay triangulation, are fundamental structures in computational geometry that partition space based on proximity to a set of seed points~\cite{aurenhammer1991voronoi,boots1999spatial}. 
Let \(P = \{\mathbf{p}_i\}_{i=1}^N\) denote a set of sites in \(\mathbb{R}^3\). These sites induce a Voronoi partition of space, where each cell
\[
V_i = \{\mathbf{x} \in \mathbb{R}^3 \mid \|\mathbf{x} - \mathbf{p}_i\| \leq \|\mathbf{x} - \mathbf{p}_j\|,\ \forall j \neq i \}
\]
contains all points closer to site \(\mathbf{p}_i\) than to any other site. The faces between neighboring cells form an explicit convex polytope structure. The list of neighbors that share a face forms an adjacency graph, namely the Delaunay graph of the sites.

\begin{figure}[t]
  \centering
  \includegraphics[width=\columnwidth]{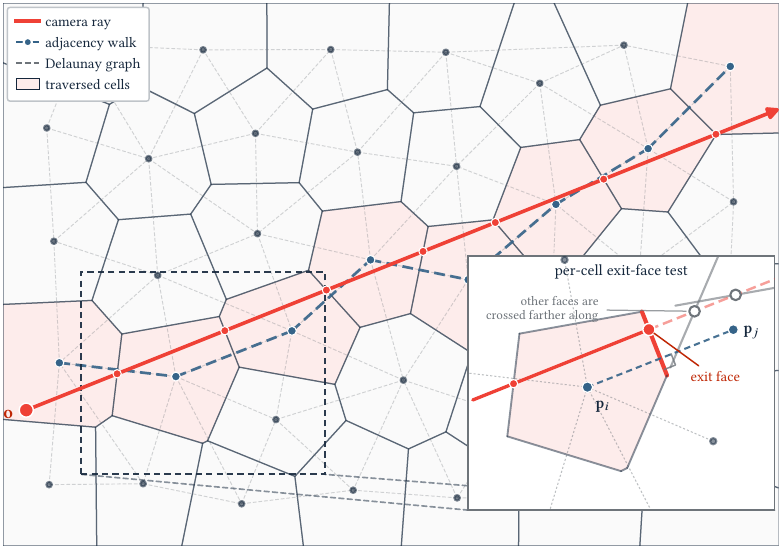}
  \caption{\textbf{Adjacency-based ray traversal.} A camera ray
    $\mathbf{r}(t)=\mathbf{o}+t\boldsymbol{\omega}$ walks a 2D Voronoi diagram
    cell by cell. Starting from the cell containing $\mathbf{o}$ (found by a
    single nearest-neighbor query), each step intersects the ray with the
    perpendicular bisector between the current site $\mathbf{p}_i$ and each of
    its Delaunay neighbors, and advances across the \emph{nearest forward}
    crossing---the cell's exit face---into the adjacent cell $\mathbf{p}_j$.
    The exit is the bisector the ray reaches first,
    while the other faces are crossed farther along. The
    traversal thus follows edges of the Delaunay adjacency graph (dashed),
    visiting a short sequence of cells whose ray-segment lengths feed the
    piecewise volume rendering.}
  \label{fig:adjacency_walk}
  \Description{Diagram of a ray crossing a two-dimensional Voronoi diagram cell by cell, with the traversed cells, Delaunay adjacency graph, and nearest exit-face test highlighted.}
\end{figure}

\paragraph{Adjacency-Based Traversal.}
Given the cell containing the ray origin, traversal proceeds by testing the ray against the bisector planes between the current site and its neighbors, selecting the nearest forward crossing, and advancing into the adjacent cell, as illustrated in \cref{fig:adjacency_walk}. The initial cell can be found with a single nearest-neighbor query. This traversal method, introduced for neural rendering by Radiant Foam~\cite{govindarajan2025radiant}, replaces repeated global nearest-intersection queries with a sequence of local exit-face tests.

The cost of one step depends on the degree of the current cell rather than directly on the total number of sites. In practice the number of neighbors varies, and the total number of cells still affects memory footprint, cache behavior, diagram construction, and the initial lookup. Nevertheless, once the walk has started, render time is governed primarily by how many cells each ray visits before termination and how much work is done per visited cell.

\subsection{Piecewise Volume Rendering}

Piecewise-constant volumetric rendering assumes that the scene is partitioned into regions of constant density and color. It is a discretization of the volume rendering equation, where the integral along a ray is computed as a sum over segments of constant properties. This is the standard rendering scheme used in NeRF~\cite{mildenhall2021nerf} and Radiant Foam~\cite{govindarajan2025radiant}. Let a ray be divided into an ordered sequence of segments $k=1,\ldots,K$, and let $i_k$ denote the index of the region containing segment $k$. The opacity $\alpha_k$ contributed by segment $k$ is
\[
  \alpha_k = 1 - \exp\bigl(-\sigma_{i_k}\,\delta_k\bigr),
\]
where \(\sigma_{i_k}\) is the density of region \(i_k\) and \(\delta_k\) is the length of the segment inside that region. The color contribution of segment $k$ is given by the cell color \(\mathbf{c}_{i_k}(\mathbf{x}_k,\boldsymbol{\omega})\), evaluated at a representative point $\mathbf{x}_k$ and view direction $\boldsymbol{\omega}$. In Voronoi tracing, these regions are the Voronoi cells and the segment boundaries are the points where the ray crosses from one cell to the next. The final pixel color is then computed by front-to-back compositing
\[
  \mathbf{C}(\mathbf{r})
    = \sum_{k} T_k\,\alpha_k\,\mathbf{c}_{i_k}(\mathbf{x}_k,\boldsymbol{\omega}),
  \qquad
  T_k = \prod_{j<k}(1 - \alpha_j),
\]
where \(T_k\) is the transmittance accumulated by segment \(k\). Early termination is possible when \(T_k\) falls below a certain threshold, which is used as a stopping criterion for ray traversal.

\subsection{Spherical Harmonic Appearance}

Within each cell the appearance does not vary across space and depends only on the viewing direction, which is what lets the framework show view-dependent effects such as specular highlights and glossy reflections. In 3D Gaussian Splatting~\cite{kerbl20233d} and Radiant Foam~\cite{govindarajan2025radiant}, this directional color is stored as spherical harmonics. The color of cell \(i\) seen from a unit direction \(\boldsymbol{\omega}\) is
\[
  \mathbf{c}_i(\boldsymbol{\omega})
    = \sum_{\ell=0}^{L} \sum_{m=-\ell}^{\ell}
      \mathbf{k}_i^{\ell m}\, Y_\ell^m(\boldsymbol{\omega}),
\]
where \(Y_\ell^m\) are the real spherical harmonic basis functions of degree \(\ell\) and order \(m\), and \(\mathbf{k}_i^{\ell m} \in \mathbb{R}^3\) are per-cell RGB coefficients learned during training. A degree-\(L\) expansion has \((L+1)^2\) coefficients per color channel. Most methods stop at \(L = 3\), which corresponds to 16 coefficients per channel, or 48 values per cell. The first term \(\mathbf{k}_i^{00}\) is the flat, view-independent color, and the higher orders add finer detail as the view changes.

Spherical harmonics are compact and fast to evaluate, but they can only represent smooth changes with direction. Sharp features such as tight specular highlights or mirror-like reflections need a high degree to capture, and the number of coefficients grows with the square of \(L\). The degree is therefore kept low in practice, so view-dependent color stays smooth and misses fine angular detail. Each coefficient also affects the whole sphere at once, so the representation cannot add detail in one direction without changing all the others.

\section{Design Considerations for Fast Ray Tracing}
\label{sec:considerations}

Voronoi tracing has different bottlenecks from both rasterization and conventional BVH ray tracing. We use the following first-order proxy for the work performed by a single ray
\begin{equation}
  \label{eq:ray-work}
  W(\mathbf{r})
  \approx
  C_{\mathrm{init}}(\mathbf{r})
  + K(\mathbf{r})
  \bigl(
    C_{\mathrm{next}}
    + C_{\mathrm{appearance}}
    + C_{\mathrm{composite}}
  \bigr).
\end{equation}
Here $C_{\mathrm{init}}(\mathbf{r})$ is the cost of locating the cell containing the ray origin and $K(\mathbf{r})$ is the number of cells traversed before termination. The remaining terms are effective average costs per visited cell. In particular, $C_{\mathrm{next}}$ covers the neighbor search and exit-face tests used to select the next cell, $C_{\mathrm{appearance}}$ covers loading and evaluating its appearance, and $C_{\mathrm{composite}}$ covers alpha compositing.

The total frame workload is obtained by summing $W(\mathbf{r})$ over the set of submitted rays $\mathcal{R}$. This sum provides a first-order workload estimate but does not predict wall-clock time directly because rays are evaluated concurrently by GPU threads. As threads are organized into warps---hardware groups of $32$ threads that execute in lockstep---throughput also depends on variation in traversal count across neighboring rays, memory locality, cache reuse, and scheduling. These effects are reflected in the effective per-cell costs in \cref{eq:ray-work}.

The model separates two direct routes to higher throughput. Reducing $K(\mathbf{r})$ shortens each traversal, while reducing the per-cell terms lowers the work performed at every step. Their realized effect also depends on memory locality and coherence across concurrently evaluated rays. The remainder of this section examines these factors and the coupling between spatial detail and cell count. \Cref{sec:method} then introduces the choices motivated by this analysis.

\subsection{Traversal Cost}

In neural rendering, it is common to evaluate rendering cost in terms of the number of primitives, such as Gaussians or voxels. Rasterization-based methods must project and sort the primitives that can contribute to the image, while BVH-based ray tracers repeatedly query a global acceleration structure. Voronoi tracing changes this dependence. After the initial cell is found, the dominant traversal cost is the number of local cell-to-cell steps taken by each ray. The total number of cells still matters through memory footprint, cache behavior, construction cost, and the initial lookup, but the per-frame traversal loop is governed primarily by $K(\mathbf{r})$.

Pushing the rendering speed of Voronoi tracing therefore requires both reducing the number of cells visited and making each step cheap. Encouraging transparent free space reduces per-cell cost as the appearance evaluation can be skipped, while concentrating opaque surfaces ensures rays terminate after fewer cells, reducing $K(\mathbf{r})$.

\begin{figure}[t]
  \centering
  \includegraphics[width=0.92\columnwidth]{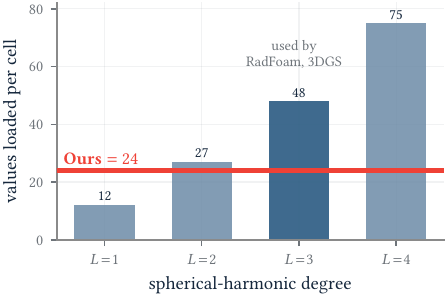}
  \caption{\textbf{Appearance values loaded per cell.} A spherical-harmonic cell
    accesses all of its RGB coefficients to evaluate color in any direction, a
    count that grows quadratically with degree ($3(L{+}1)^2$); at degree $3$, the
    setting used by Radiant Foam and 3DGS, this is $48$ values. Our deployed
    representation performs two bilinear lookups, one in each appearance map.
    Each lookup requires four RGB texels, giving a fixed total of $24$ appearance
    values independent of map resolution, halving the appearance traffic.}
  \label{fig:bytes_per_cell}
  \Description{Bar chart comparing the number of appearance values loaded per cell for spherical harmonics of increasing degree and for the proposed octahedral texture lookup.}
\end{figure}

In our profile of Radiant Foam~\cite{govindarajan2025radiant} on the Mip-NeRF~360 Garden scene, neighbor search accounts for roughly three quarters of renderer kernel time, including adjacency loads and the intersection tests that find the exit face. Color evaluation accounts for most of the remainder, and compositing is comparatively small. The kernel is also largely memory bound. Within color evaluation, most of the cost is loading spherical-harmonic coefficients. As \cref{fig:bytes_per_cell} shows, this load grows quadratically with spherical-harmonic degree, whereas the bilinear texture lookups we adopt (\cref{sec:octahedral}) access a fixed, smaller number of stored values regardless of texture resolution.

\subsection{Memory Locality}

Voronoi tracing renders each pixel by traversing and compositing an ordered sequence of cells, evaluating the appearance of each cell along the ray. Unlike tile-based splatting, which cooperatively loads the primitives shared by a screen tile and amortizes their cost across many pixels, a Voronoi traversal issues appearance loads on demand for each ray. This makes memory locality critical. Neighboring camera rays often traverse overlapping cell sequences, so grouping coherent rays can improve cache reuse even without explicit cooperative loading. Locality also depends on the cell layout in memory, since cells that are close in space are more likely to be visited by nearby rays and should ideally map to nearby memory addresses. We use these observations in the inference implementation described in \cref{sec:inference-optimizations} and ablate their effect in \cref{sec:speed_analysis}.

\subsection{Spatial Detail and Cell Count}

A spherical-harmonic cell encodes only \emph{angular} variation, since it assigns a color based only on viewing direction. High-frequency appearance is therefore expressible across views, but not within the image. Within a single rendered image, all rays that strike a given cell arrive from nearly the same direction, so SH yields an almost constant color over the cell's image footprint. Spatial texture detail can then be introduced only by subdividing the space into additional cells, coupling spatial detail directly to cell count.

This coupling has two undesirable consequences. First, the number of cells must increase rapidly with the desired level of detail, and we observe diminishing returns together with increasingly visible aliasing as the number of sites grows. Second, and more consequential for optimization, each cell is observed by progressively fewer rays as cells shrink. In the limit, a cell is seen by a single ray from a single direction and overfits to that observation. This is detrimental because recovering 3D shape from images without depth supervision relies on each cell being constrained by many rays from diverse viewpoints. Once a cell is coupled to a single ray, the geometry is under-constrained and is free to settle into a degenerate configuration.

Radiant Foam mitigates these effects through optimization heuristics. 
It begins training at lower resolution images and with fewer initial points, on the order of $100{,}000$ points, so that coarse geometry is recovered before the representation can overfit to high frequency detail. Throughout the training, the resolution of training images increases, the model densifies the number of cells and prunes unused cells, reaching up to $4M$ cells.

\section{Method}
\label{sec:method}

Based on these design considerations, we propose \methodname{}, a differentiable renderer for an explicit Voronoi scene representation. We reduce per-cell appearance traffic using octahedral surface and view-dependent textures (\cref{sec:octahedral,sec:view-depend}), while surface-concentrated opacity and a scale-invariant density parameterization shorten ray traversal (\cref{sec:surface-opacity,sec:scale-inv-dens}). We optimize a fixed budget of densely initialized cells without pruning or densification (\cref{sec:dense-init-and-training}) and use a locality-aware GPU implementation at inference (\cref{sec:inference-optimizations}).

\subsection{Octahedral Surface Texture}
\label{sec:octahedral}

The appearance formulation is where our method departs most clearly from prior Voronoi-based rendering.
We replace per-cell spherical harmonics with a small per-cell texture that is projected onto the cell surface, allowing a single cell to exhibit spatial appearance variation within one view. Concretely, we assign to each site $\mathbf{p}_i$ an $8\times8$ RGB texture $T_i$. A Voronoi cell is an intersection of half-spaces and is therefore convex, with its site contained in its interior. Consequently, every ray cast from the site exits through exactly one point of the cell boundary, and the map from directions on the sphere $S^2$ to surface points is a bijection. This lets us parameterize a cell's appearance by direction from the site. When a camera ray intersects the surface of cell $V_i$ at a point $\mathbf{x}$, we form the unit direction
\[
  \mathbf{d} = \frac{\mathbf{x} - \mathbf{p}_i}
                    {\lVert \mathbf{x} - \mathbf{p}_i \rVert} \in S^2 ,
\]
and use $\mathbf{d}$ to index $T_i$.

\begin{figure}[t]
  \centering
  \includegraphics[width=\columnwidth]{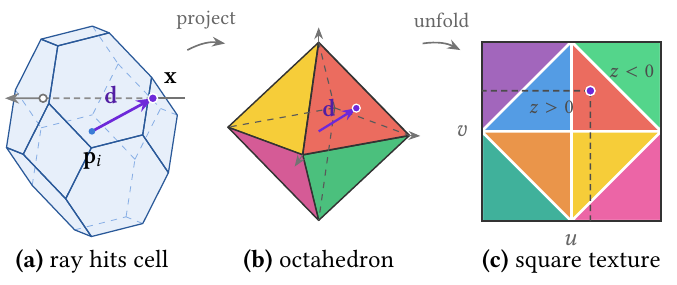}
  \caption{%
    \textbf{Octahedral appearance lookup.}
    \textbf{(a)} A camera ray crosses a Voronoi cell; the unit direction
    $\mathbf{d}$ from the site $\mathbf{p}_i$ to the hit point (violet) selects
    the appearance.
    \textbf{(b)} $\mathbf{d}$ is $\ell_1$-normalized onto the unit octahedron,
    landing on one of its eight faces (here the red octant);
    the eight faces are color-coded by octant.
    \textbf{(c)} The octahedron is unfolded into the square texture domain, where the
    upper hemisphere ($d_z\!\ge\!0$) fills the central diamond and the lower
    hemisphere ($d_z\!<\!0$) the corners.
  }
  \label{fig:octahedral}
  \Description{Three-panel diagram. Left: a camera ray strikes a convex Voronoi
    cell at a point, and the unit direction from the cell's site to that point
    is highlighted. Middle: a 3D octahedron whose eight faces are color-coded
    by octant, with the highlighted direction landing on the red face. Right:
    the octahedron unfolded into a square, its eight faces shown as colored
    triangles, with the looked-up texel marked in the matching red wedge.}
\end{figure}

We use octahedral mapping to index from the direction to the square texture domain. This provides low distortion, and is free of the polar singularities of latitude--longitude maps, remaining continuous across the texture boundary. The direction $\mathbf{d}$ is first projected onto the unit octahedron by $\ell_1$ normalization,
\[
  \mathbf{q} = \frac{\mathbf{d}}{\lvert d_x\rvert + \lvert d_y\rvert + \lvert d_z\rvert}.
\]
The upper hemisphere maps directly to the square texture, while the lower hemisphere is unfolded to the surrounding region.
\[
  (u,v) =
  \begin{cases}
    (q_x,\; q_y), & d_z \ge 0,\\[4pt]
    \bigl((1 - \lvert q_y\rvert)\,\mathrm{sign}(q_x),\;
          (1 - \lvert q_x\rvert)\,\mathrm{sign}(q_y)\bigr), & d_z < 0.
  \end{cases}
\]
These coordinates $(u,v) \in [-1,1]^2$ are then rescaled to texel space. Bilinear interpolation provides smoothly varying color from the $8\times8$ texture. \Cref{fig:octahedral} illustrates this mapping from a cell sample direction to the square texture.

\subsection{View-Dependent Texture}
\label{sec:view-depend}

The surface texture $T^{\text{vi}}_i$ introduced above is, by design, independent of the camera. A surface point returns the same color from any viewpoint. Real scenes, however, have variable color dependent on viewing direction, including specular highlights, sheen, and reflections of the surrounding environment. We capture these effects with a second per-cell texture $T^{\text{vd}}_i$ of the same form, but indexed by the view direction rather than by the surface intersection point. The two maps combine additively in logit space, and a sigmoid produces the final cell color,
\[
  \mathbf{c}_i(\mathbf{x}, \boldsymbol{\omega})
  = \sigma\!\bigl(T^{\text{vi}}_i(\mathrm{oct}(\mathbf{d}))
                + T^{\text{vd}}_i(\mathrm{oct}(\boldsymbol{\omega}))\bigr),
\]
where $\boldsymbol{\omega}$ corresponds to the unit vector from the hit point towards the camera. 

The two maps separate two complementary sources of appearance variation. Within a single image, $T^{\text{vi}}_i$ varies across the cell's footprint as $\mathbf{d}$ changes from pixel to pixel, while $T^{\text{vd}}_i$ is effectively constant. Across different views of the scene the roles reverse. $T^{\text{vi}}_i$ returns the same color at each surface point, while $T^{\text{vd}}_i$ varies with the moving camera.

In practice we regularize the view-dependent term to remain small relative to $T^{\text{vi}}_i$, so that it acts as a residual on top of the surface texture. The precise penalty is given in \cref{sec:dense-init-and-training}.

\begin{figure*}[t]
  \centering
  \includegraphics[width=\textwidth]{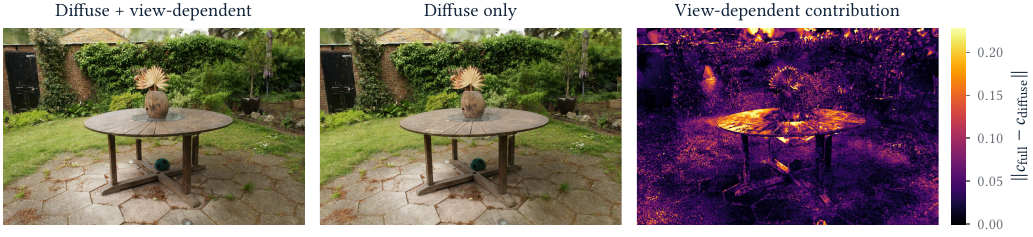}
  \caption{\textbf{View-dependent contribution.} We render a held-out view with
    the full model and again with the view-dependent map $T^{\text{vd}}$ zeroed
    (a pure-diffuse render); density is unchanged, so the difference isolates the
    view-dependent term. Diffuse regions remain similar, while glossy and
    reflective surfaces show clear changes. The right panel shows that the
    residual is localized on reflective tabletop objects and foliage, exactly
    the surfaces whose appearance changes with viewpoint.}
  \label{fig:specular_map}
  \Description{Wide figure comparing the full render, the diffuse-only render with the view-dependent effects disabled, and a difference visualization that highlights reflective objects and foliage.}
\end{figure*}

\Cref{fig:specular_map} isolates this term by rendering a view with and without the view-dependent map. Diffuse regions change little, while glossy and reflective surfaces show clear differences.

\subsection{Surface-Concentrated Opacity and Geometry}
\label{sec:surface-opacity}

\begin{figure}[t]
  \centering
  \includegraphics[width=0.92\columnwidth]{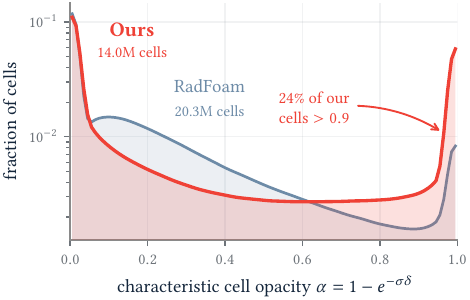}
  \caption{\textbf{Cell-opacity distribution on the Mip-NeRF~360 scenes.} Characteristic
    per-cell opacity $\alpha = 1-e^{-\sigma\delta}$, where $\delta$ is the mean
    distance to a cell's Voronoi neighbors. Our exponential density
    parameterization and distortion loss drives opacity to a strongly bimodal distribution --- cells
    are either near-transparent or near-opaque --- with $24\%$ of cells above
    $\alpha=0.9$. Radiant Foam's softplus model instead leaves most cells
    semi-transparent ($44\%$ in $[0.1,0.9]$, only $4\%$ above $0.9$), so rays must
    composite many partially-occupied cells before terminating. Our method
    reaches this with roughly half the cells ($2.0$M vs.\ $4.1$M).}
  \label{fig:opacity_hist}
  \Description{Histogram comparing per-cell opacity distributions; our method has peaks near transparent and opaque cells, while Radiant Foam has more semi-transparent cells.}
\end{figure}

As previously mentioned, computation cost is mostly tied to the number of traversed cells. For the mostly opaque scenes considered here, density should converge toward a thin, fully opaque surface rather than a cloud of semi-transparent cells. Ideally, a ray traverses empty cells until it reaches one opaque cell whose surface texture explains the visible color. Concentrating opacity in this way reduces both compositing work and the number of expensive appearance evaluations. \Cref{fig:opacity_hist} shows that our model drives per-cell opacity to a strongly bimodal distribution, in contrast to the semi-transparent cells that dominate Radiant Foam.

To push the optimization close to an opaque surface, we adopt the distortion loss of Mip-NeRF~360~\cite{barron2022mip}. For a ray with compositing weights $w_k = T_k\alpha_k$ over traversed intervals $[t_k,t_{k+1}]$, metric depth is first mapped to the bounded disparity-like coordinate
\[
  s(t) = 1 - \frac{1}{1+t}.
\]
Let $\bar{s}_k = \frac{1}{2}(s(t_k)+s(t_{k+1}))$ and $\Delta s_k = s(t_{k+1})-s(t_k)$. The per-ray distortion penalty is
\begin{equation}
  \label{eq:distortion-loss}
  \mathcal{L}_{\mathrm{dist}}(\mathbf{r})
  =
  \sum_{k,\ell} w_k w_\ell
  \left\lvert \bar{s}_k - \bar{s}_\ell \right\rvert
  +
  \frac{1}{3}\sum_k w_k^2 \Delta s_k .
\end{equation}
The first term penalizes pairs of compositing weights at separated depths, while the second penalizes the extent of intervals carrying compositing weight. Unlike a depth supervision term, this loss does not prescribe where the surface should lie; it only encourages the rendering weights that already explain the image to become compact. This is well aligned with fast Voronoi tracing because compact weights imply transparent free-space cells followed by early termination at an opaque surface cell.

\subsection{Scale-Invariant Density Parameterization}
\label{sec:scale-inv-dens}
The Voronoi partition produces cells of widely varying size, which is desirable, as it concentrates resolution where the scene demands it. Radiant Foam assigns each cell a softplus-activated constant density $\sigma_i$ and converts it to opacity over a ray segment of length $\delta_i$ as $\alpha_i=1-\exp(-\sigma_i\delta_i)$. Because smaller cells generally produce shorter ray segments, they require a larger density to reach the same opacity. Precisely, the density required for a given opacity is
\[
\sigma_i = -\ln(1-\alpha_i)/\delta_i
\]
which scales \emph{inversely} with $\delta_i$.
On the other hand, differentiating the opacity with respect to density gives
\begin{equation}
  \label{eq:density-grad}
  \frac{\partial L}{\partial \sigma_i}
    = \frac{\partial L}{\partial \alpha_i}\,\delta_i\,(1-\alpha_i),
\end{equation}
which scales \emph{linearly} with the segment length $\delta_i$. A smaller cell therefore requires a higher density to represent the same opacity, while the gradient with respect to its density is proportionally weaker.

In practice this manifests as floaters and a persistent haze, where large
cells that should be empty settle at a low but non-zero density around
surfaces. The effect is most severe when training from a dense initialization, where it can drive a runaway increase in the
density of large cells in free space near the camera. Once opaque, early ray termination starves the cells behind them of gradient and the optimization cannot recover.

The softplus density activation used in prior works~\cite{govindarajan2025radiant,govindarajan2026power} does not address this issue. We instead set $\sigma = \exp(\rho)$, whose derivative satisfies $\partial\sigma/\partial\rho = \sigma$. The gradient on the optimized parameter becomes
\begin{equation}
  \label{eq:rho-grad}
  \frac{\partial L}{\partial \rho}
    = \frac{\partial L}{\partial \sigma_i}\,\sigma_i
    = \frac{\partial L}{\partial \alpha_i}\,\delta_i\,(1-\alpha_i)\,\sigma_i
    = \frac{\partial L}{\partial \alpha_i}\,(1-\alpha_i)\,
      \ln\!\frac{1}{1-\alpha_i},
\end{equation}
where the final equality is the identity $\sigma_i\delta_i = -\ln(1-\alpha_i)$ and the segment length cancels. Two cells of equal opacity thus receive an identically scaled gradient regardless of their size, and optimization treats small and large cells on equal footing, removing the systematic bias against finely tiled surface detail.

\subsection{Simplified Heuristic-Free Training}
\label{sec:dense-init-and-training}

Adaptive density control can make a representation robust to sparse or uneven structure-from-motion points, and better strategies for adding and removing cells remain an interesting direction. Our goal here is instead to isolate how much quality and speed can be obtained from a fixed, well-initialized Voronoi representation. In Radiant Foam~\cite{govindarajan2025radiant}, pruning and densification decide where cells should be removed or added during training. We found this difficult to make reliable, since it can overpopulate regions where extra expressiveness brings little benefit while still missing under-resolved surfaces. We therefore use a fixed cell budget, where all sites, densities, and textures are optimized directly after initialization and no cells are inserted or removed.

Following the dense-initialization strategy explored by EDGS~\cite{kotovenko2026edgs}, we generate a dense point cloud once at the beginning of training. For each scene, we select up to $100$ representative reference images by clustering camera poses. Each reference image is matched against its $3$ nearest neighboring cameras using RoMa v2~\cite{edstedt2025roma}. From every image pair, we sample $15{,}000$ RoMa correspondences according to matching confidence, triangulate them with the calibrated poses and intrinsics, and reject points with negative depth, non-finite coordinates, or reprojection error above $2$ pixels. The color of each accepted point is taken from the reference image.

The resulting point cloud is subsampled to the training budget using density-aware sampling. We estimate local sampling density with a $128^3$ voxel grid and draw points with probability proportional to the inverse voxel occupancy, which avoids spending the entire budget in over-matched regions. We additionally add a small number of random background points around the cloud to cover empty space. In our experiments we train with a fixed budget of $2$M sites, of which $5{,}000$ are random background points. The corresponding Voronoi cells and adjacency are constructed on the GPU using the scalable algorithm of \citet{taveira2026paragram}. The diffuse texture of each initialized point is set by repeating the logit of its RGB color across all texels, the view-dependent texture is initialized to zero, and the density is initialized low for random background points.

We then optimize for $20{,}000$ steps using Adam over site positions, density parameters, and both texture maps. Training uses the same image resolution and the same fixed cell set from the first iteration to the last, without progressive downsampling, pruning, densification, or multi-stage schedules. The objective contains only photometric reconstruction and regularization,
\begin{equation}
  \label{eq:training-loss}
  \mathcal{L}
  =
  \mathcal{L}_{\mathrm{rgb}}
  + \lambda_{\mathrm{dist}}\mathcal{L}_{\mathrm{dist}}
  + \lambda_{\mathrm{vd}}\mathcal{L}_{\mathrm{vd}}
  + \lambda_{\mathrm{mean}}\mathcal{L}_{\mathrm{mean}} .
\end{equation}
The Smooth-$L_1$ term $\mathcal{L}_{\mathrm{rgb}}$ matches the composited rendered color to the training image. The distortion term $\mathcal{L}_{\mathrm{dist}}$ from \cref{eq:distortion-loss} concentrates opacity along each ray. The view-dependent regularizer
\[
  \mathcal{L}_{\mathrm{vd}}
  =
  \frac{1}{N R_{\mathrm{vd}}^2}
  \sum_i \left\lVert T_i^{\mathrm{vd}} \right\rVert_2^2
\]
keeps specular appearance as a small residual on top of the view-independent texture. Finally, since many texels of a cell may be unobserved from the training cameras, we pull diffuse texels toward their per-cell mean,
\[
  \mathcal{L}_{\mathrm{mean}}
  =
  \frac{1}{N R_{\mathrm{vi}}^2}
  \sum_{i,u}
  \left\lVert
    T_{i,u}^{\mathrm{vi}}
    - \frac{1}{R_{\mathrm{vi}}^2}\sum_v T_{i,v}^{\mathrm{vi}}
  \right\rVert_2^2 .
\]
This regularizer propagates the observed cell color to unseen directions without forcing neighboring cells or unrelated surface regions to be smooth.

\subsection{Inference Optimizations}
\label{sec:inference-optimizations}
The training formulation above is already designed to reduce the number of cells visited per ray. Our base inference kernel follows Radiant Foam~\cite{govindarajan2025radiant} in using half-precision appearance attributes and precomputed half-precision vectors from each site to its Voronoi neighbors. The latter avoid loading two point positions and computing their difference during traversal. For our texture representation, each RGB texel is padded to four channels. Although this adds one unused value per texel, it lets the CUDA kernel read a complete texel with a single aligned load.

We additionally reorder cells by Morton code before rendering, improving spatial locality for rays that traverse neighboring regions of the diagram.

To improve coherence within a warp, we schedule image rays in compact $4\times8$ tiles. The output is still written in image order, but neighboring threads tend to traverse similar cells and therefore share cache lines. For cells whose compositing weight is below a small threshold, the inference kernel updates transmittance but skips the texture lookup and sigmoid evaluation, avoiding appearance loads that cannot affect the final color appreciably. This threshold exposes an explicit speed-quality trade-off, which we ablate in \cref{sec:speed_analysis}.

\section{Results}

We evaluate our method along three axes. First, \cref{sec:quantitative_evaluation} reports reconstruction quality and rendering speed on Mip-NeRF~360 for ray-based and rasterized NVS methods, while \cref{sec:qualitative_comparison} examines representative visual differences. Second, \cref{sec:speed_analysis,sec:ablations} analyze where the speedup comes from by measuring traversal cost, cells per ray, and inference-time approximations. Third, \cref{sec:applications} demonstrates how the ray-based renderer supports non-pinhole camera effects through ray generation without changes to a rasterization pipeline. Together, these experiments test whether a carefully designed ray-based representation can retain ray-tracing flexibility while achieving real-time throughput.

\subsection{Quantitative Evaluation}
\label{sec:quantitative_evaluation}

We evaluate on the Mip-NeRF~360 dataset~\cite{barron2022mip}, which contains outdoor and indoor object-centric scenes reconstructed with COLMAP~\cite{schoenberger2016sfm, schoenberger2016mvs}. We use the standard train-test split, holding out every eighth image for evaluation, and report averages separately for outdoor scenes, indoor scenes, and the full dataset. All images are evaluated at the same dataset resolution used by the corresponding methods, $4\times$ downsampling for outdoor scenes and $2\times$ for indoor. We train each scene on a single NVIDIA GeForce RTX~5090. The $20{,}000$-step optimization takes $33$--$50$ minutes per scene, averaging $40$ minutes, including the one-time correspondence-based point-cloud construction.

Reconstruction quality is measured with PSNR, SSIM, and LPIPS on held-out views using one metric implementation shared by all methods. We retain the background convention of each published configuration. Rendering speed is measured on the same GPU and reported in frames per second for full-image rendering after model loading and one-time, view-independent preprocessing. The timing protocol follows the use case of an interactive viewer, where each requested frame is rendered end-to-end without precomputations that depend on knowing future camera poses. For our method, the timed region includes the initial cell search, Voronoi traversal, appearance evaluation, and compositing. It excludes only scene-level preprocessing that would be performed once after loading the model, such as conversion to half-precision storage and Morton reordering of cells. When rerunning baselines, we apply the same viewer-style rule. Methods may perform model loading and scene-level setup, but not precompute quantities that depend on the sequence of future test cameras. Further details on the evaluation protocol and baseline provenance are provided in the supplementary material.

We measure runtime with CUDA events on held-out images. Each view is warmed up before timing, then rendered repeatedly with the same code path used for normal evaluation, and the per-frame time is averaged over the test views. Our reported numbers use 10 warmup renders and 20 timed renders per view, which measures steady-state full-frame rendering while keeping all per-frame work inside the timed region.

We compare against two groups of methods. The ray-based comparison includes iNGP~\cite{muller2022instant}, Zip-NeRF~\cite{barron2023zip}, 3DGRT~\cite{moenne20243d}, Radiant Foam~\cite{govindarajan2025radiant}, and the ray-traced rendering mode of PowerFoam~\cite{govindarajan2026power}. PowerFoam is trained in a rasterized pipeline but can be ray-traced in inference. Zip-NeRF is the only method in this comparison not trained on the RTX~5090; we train it with the official Zip-NeRF implementation on a $4\times$ GH200 setup and report an approximate RTX~5090 rendering rate measured with a compatible JAX runtime. The rasterized comparison includes 3DGS~\cite{kerbl20233d}, 3DGUT~\cite{wu20253dgut}, Radiance Meshes~\cite{mai2026radiance}, Triangle Splatting~\cite{held2026triangle}, and PowerFoam, providing the broader real-time NVS context. The comparisons use the same dataset split, image resolutions, metric implementation, and viewer-style timing boundary. Method-specific timing interfaces and deviations are documented in the supplementary material. For Radiance Meshes, the high-throughput Vulkan renderer did not reproduce the quality of the training pipeline in our tests, so we report the quality metrics and rendering speed of the Python/CUDA and Vulkan implementations separately.

\definecolor{tbf}{rgb}{1, 0.70, 0.70} %
\definecolor{tbs}{rgb}{1, 0.85, 0.70} %
\definecolor{tbt}{rgb}{1, 1.00, 0.70} %
\definecolor{tbn}{rgb}{1, 1.00, 1.00} %

\begin{table*}[t]
    \small
    \centering
    \setlength{\tabcolsep}{3pt}
    \caption{\textbf{Mip-NeRF~360, ray-based methods.} Per metric column,
    \colorbox{tbf}{best}, \colorbox{tbs}{second} and \colorbox{tbt}{third}.
    \methodname{} renders $2$--$3\times$ faster than every other ray-based method while
    improving on its predecessor Radiant Foam across all quality metrics.}
    \label{tab:mip360_raytraced}
    \begin{tabular}{@{}l*{4}{c}!{\color{black!35}\vrule width 0.6pt}*{4}{c}!{\color{black!35}\vrule width 0.6pt}*{4}{c}@{}}
    \toprule
    Method &
        \multicolumn{4}{c}{Outdoor} &
        \multicolumn{4}{c}{Indoor} &
        \multicolumn{4}{c}{All} \\
    \cmidrule(lr){2-5}\cmidrule(lr){6-9}\cmidrule(lr){10-13}
     & PSNR$\uparrow$ & SSIM$\uparrow$ & LPIPS$\downarrow$ & FPS$\uparrow$
     & PSNR$\uparrow$ & SSIM$\uparrow$ & LPIPS$\downarrow$ & FPS$\uparrow$
     & PSNR$\uparrow$ & SSIM$\uparrow$ & LPIPS$\downarrow$ & FPS$\uparrow$ \\
    \midrule
    Zip-NeRF &
        \cellcolor{tbf}27.13 & \cellcolor{tbf}0.808 & \cellcolor{tbf}0.197 & 0.29 &
        \cellcolor{tbf}32.29 & \cellcolor{tbf}0.929 & \cellcolor{tbf}0.197 & 0.19 &
        \cellcolor{tbf}30.08 & \cellcolor{tbf}0.877 & \cellcolor{tbf}0.197 & 0.23 \\
    iNGP (base) &
        25.16 & 0.636 & 0.397 & 30 &
        28.59 & 0.841 & 0.340 & 25 &
        27.12 & 0.753 & 0.365 & 27 \\
    iNGP (big) &
        \cellcolor{tbs}26.01 & 0.697 & 0.342 & 28 &
        29.45 & 0.863 & 0.297 & 24 &
        27.98 & 0.792 & 0.316 & 26 \\
    3DGRT &
        \cellcolor{tbt}25.90 & \cellcolor{tbs}0.785 & \cellcolor{tbs}0.219 & 103 &
        29.93 & \cellcolor{tbt}0.914 & \cellcolor{tbt}0.240 & 77 &
        28.20 & \cellcolor{tbs}0.859 & \cellcolor{tbs}0.231 & 88 \\
    RadFoam &
        25.40 & 0.731 & 0.298 & \cellcolor{tbs}156 &
        \cellcolor{tbt}30.72 & 0.902 & 0.261 & \cellcolor{tbs}222 &
        28.44 & 0.829 & 0.277 & \cellcolor{tbs}194 \\
    PowerFoam (RT) &
        25.39 & 0.730 & \cellcolor{tbt}0.278 & \cellcolor{tbt}124 &
        \cellcolor{tbt}31.31 & 0.913 & 0.241 & \cellcolor{tbt}136 &
        \cellcolor{tbt}28.78 & 0.835 & 0.257 & \cellcolor{tbt}131 \\
    \textbf{\methodname{}} &
        25.74 & \cellcolor{tbt}0.757 & \cellcolor{tbt}0.252 & \cellcolor{tbf}678 &
        \cellcolor{tbs}31.42 & \cellcolor{tbs}0.916 & \cellcolor{tbs}0.222 & \cellcolor{tbf}582 &
        \cellcolor{tbs}28.98 & \cellcolor{tbt}0.848 & \cellcolor{tbt}0.235 & \cellcolor{tbf}623 \\
    \bottomrule
    \end{tabular}
\end{table*}

\begin{table*}[t]
    \small
    \centering
    \setlength{\tabcolsep}{3pt}
    \caption{\textbf{Mip-NeRF~360, rasterized methods vs.\ ours.} Per metric column,
    \colorbox{tbf}{best}, \colorbox{tbs}{second} and \colorbox{tbt}{third}.
    \methodname{} is the only ray-traced method here; it trails the strongest
    rasterizers slightly on appearance metrics yet renders roughly $2\times$ faster
    than all of them.}
    \label{tab:mip360_rasterized}
    \begin{tabular}{@{}l*{4}{c}!{\color{black!35}\vrule width 0.6pt}*{4}{c}!{\color{black!35}\vrule width 0.6pt}*{4}{c}@{}}
    \toprule
    Method &
        \multicolumn{4}{c}{Outdoor} &
        \multicolumn{4}{c}{Indoor} &
        \multicolumn{4}{c}{All} \\
    \cmidrule(lr){2-5}\cmidrule(lr){6-9}\cmidrule(lr){10-13}
     & PSNR$\uparrow$ & SSIM$\uparrow$ & LPIPS$\downarrow$ & FPS$\uparrow$
     & PSNR$\uparrow$ & SSIM$\uparrow$ & LPIPS$\downarrow$ & FPS$\uparrow$
     & PSNR$\uparrow$ & SSIM$\uparrow$ & LPIPS$\downarrow$ & FPS$\uparrow$ \\
    \midrule
    3DGS &
        \cellcolor{tbf}26.45 & \cellcolor{tbf}0.801 & \cellcolor{tbs}0.204 & 143 &
        \cellcolor{tbt}31.11 & \cellcolor{tbt}0.924 & 0.237 & 279 &
        \cellcolor{tbf}29.11 & \cellcolor{tbf}0.872 & \cellcolor{tbs}0.223 & 220 \\
    3DGUT &
        \cellcolor{tbs}26.21 & \cellcolor{tbs}0.796 & \cellcolor{tbt}0.211 & 183 &
        31.07 & \cellcolor{tbs}0.925 & \cellcolor{tbt}0.235 & 307 &
        \cellcolor{tbs}28.98 & \cellcolor{tbs}0.870 & \cellcolor{tbt}0.225 & 254 \\
    PowerFoam (rast.) &
        25.39 & 0.730 & 0.278 & \cellcolor{tbt}222 &
        \cellcolor{tbs}31.31 & 0.913 & 0.241 & \cellcolor{tbs}350 &
        28.78 & 0.835 & 0.257 & \cellcolor{tbt}295 \\
    Radiance Meshes &
        \cellcolor{tbt}26.11 & 0.783 & 0.249 & 59 &
        30.16 & 0.919 & 0.240 & 50 &
        28.43 & 0.861 & 0.244 & 53 \\
    Radiance Meshes (Vulkan) &
        23.98 & 0.649 & 0.318 & \cellcolor{tbs}317 &
        28.07 & 0.847 & 0.274 & \cellcolor{tbt}345 &
        26.32 & 0.762 & 0.293 & \cellcolor{tbs}333 \\
    Triangle Splatting &
        26.03 & \cellcolor{tbt}0.793 & \cellcolor{tbf}0.196 & 115 &
        30.83 & \cellcolor{tbf}0.925 & \cellcolor{tbf}0.205 & 179 &
        28.78 & \cellcolor{tbt}0.869 & \cellcolor{tbf}0.201 & 151 \\
    \textbf{\methodname{}} &
        25.74 & 0.757 & 0.252 & \cellcolor{tbf}678 &
        \cellcolor{tbf}31.42 & 0.916 & \cellcolor{tbs}0.222 & \cellcolor{tbf}582 &
        \cellcolor{tbt}28.98 & 0.848 & 0.235 & \cellcolor{tbf}623 \\
    \bottomrule
    \end{tabular}
\end{table*}

\Cref{tab:mip360_raytraced} compares against methods that cast rays through a 3D representation. Our method reaches 623 FPS on average, compared with 194 FPS for Radiant Foam and 131 FPS for the ray-traced PowerFoam, while improving over both in PSNR, SSIM, and LPIPS. The quality gain is largest on indoor scenes, where the octahedral texture representation and dense initialization improve high-frequency appearance without increasing traversal cost.

\Cref{tab:mip360_rasterized} places the same results next to widely used rasterized real-time NVS methods. Our method is $2.8\times$ faster than 3DGS on average and reaches the best indoor PSNR in the table. The strongest rasterizers still lead on outdoor appearance metrics, where 3DGS is $0.71$\,dB higher in PSNR and Triangle Splatting gives the best LPIPS. The main point of this comparison is not that ray tracing is universally faster than rasterization, but that a ray-based representation can remain competitive with---and in this benchmark exceed the throughput of---the rasterized methods most commonly used in current research and practice.

Unlike the compared baselines, our method does not perform densification during training. This keeps the cell set fixed and simplifies the training procedure, but it also limits the model's ability to allocate new capacity to regions that remain poorly reconstructed. The effect is most visible in outdoor scenes with dense foliage, distant structure, and high-frequency texture. As a result, our final quality depends more strongly on the dense initialization than on adaptive growth during optimization.

\subsection{Qualitative Comparison}
\label{sec:qualitative_comparison}

\begin{figure*}[!t]
  \centering
  \includegraphics[width=\textwidth,height=0.68\textheight,keepaspectratio]{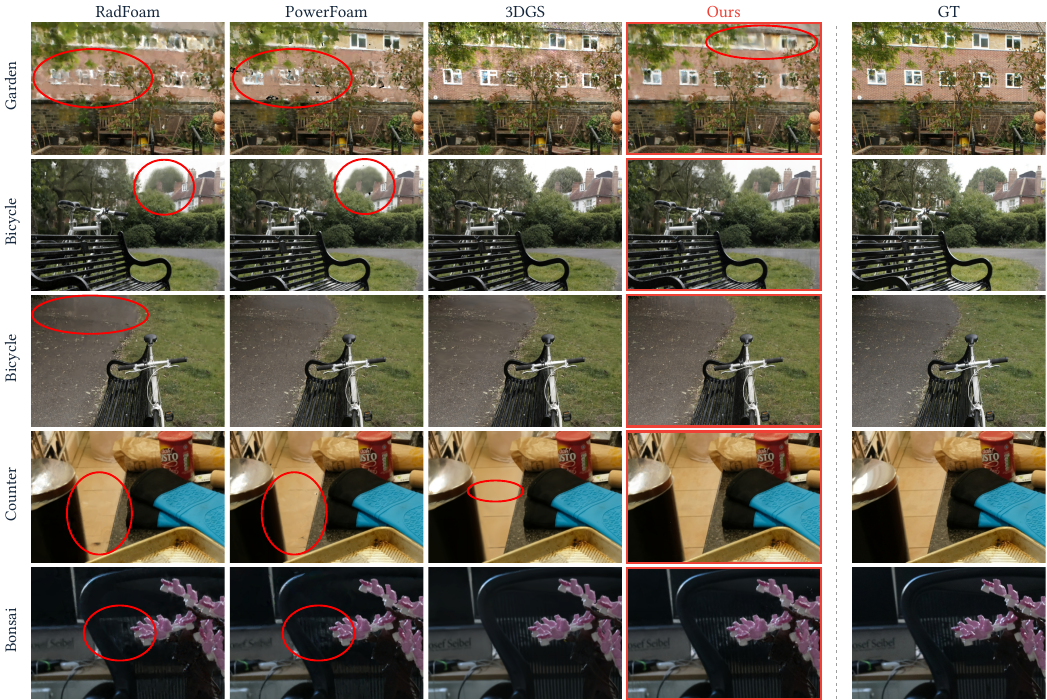}
  \caption{\textbf{Cropped qualitative comparison on held-out Mip-NeRF~360 views.}
    We show challenging regions that are observed by few training images and
    therefore stress fine-scale reconstruction. \methodname{} uses a fixed
    budget of $2$M cells, half the Radiant Foam budget on outdoor scenes. Unlike
    methods that densify during training, it cannot add cells later to recover
    underrepresented regions; this limitation is visible in some crops, where
    \methodname{} preserves more structure than Radiant Foam but remains less
    detailed than 3DGS. Red circles mark representative artifacts or missing details in the compared reconstructions.}
  \label{fig:qualitative_multimethod}
  \Description{Grid of cropped held-out Mip-NeRF 360 views comparing Radiant Foam, PowerFoam, 3DGS, our method, and ground truth across garden, bicycle, counter, and bonsai scenes.}
\end{figure*}

\Cref{fig:qualitative_multimethod} shows zoomed crops from challenging held-out views, focusing on high-frequency background regions that are observed by relatively few training images. In these cases, \methodname{} often preserves more local structure than Radiant Foam and PowerFoam, for example in the kitchen floor tiles in Counter and the textile pattern on the back of the chair in Bonsai. The fixed cell budget remains a limitation, however. In Garden, where fine foliage and distant brick texture require highly localized capacity, 3DGS preserves sharper detail, suggesting that adaptive density control is still important in underrepresented regions.

\subsection{Rendering Speed}
\label{sec:speed_analysis}

\begin{figure}[t]
  \centering
  \includegraphics[width=0.92\columnwidth]{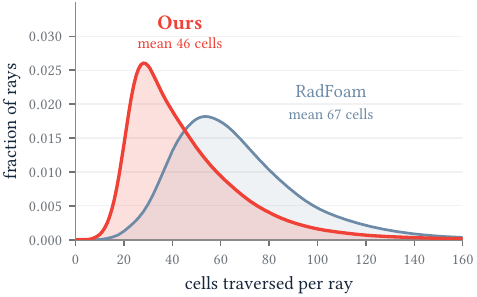}
  \caption{\textbf{Cells traversed per ray}, ours vs.\ Radiant Foam on
    held-out views of the Mip-NeRF~360 scenes. Unlike rasterization or BVH ray tracing,
    the cost of Voronoi traversal is set by this quantity rather than the total
    cell count $N$. Our surface-concentrated opacity (\cref{fig:opacity_hist}) shifts the whole
    distribution left. Rays visit $46.1$ cells on average versus
    $66.9$ for Radiant Foam, a $31\%$ reduction in mean traversal count, while we also reach
    higher reconstruction quality (\cref{tab:mip360_raytraced}).}
  \label{fig:cells_per_ray}
  \Description{Distribution plot showing that our method shifts the cells-per-ray histogram left relative to Radiant Foam, with a lower mean traversal count.}
\end{figure}

\begin{figure*}[t]
  \centering
  \includegraphics[width=\textwidth]{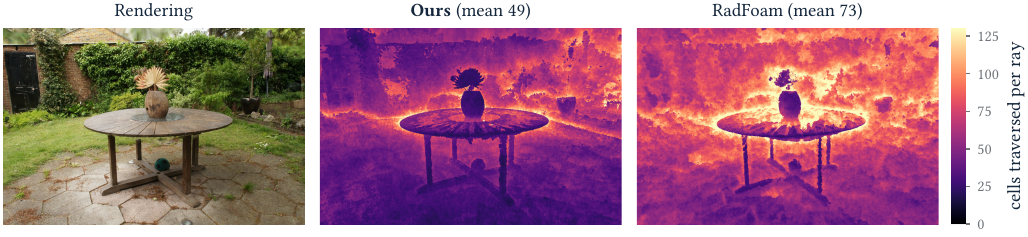}
  \caption{\textbf{Cells traversed per ray, in image space.} For a held-out view
    of the garden scene we map the per-pixel number of cells each ray visits
    before terminating. Cost concentrates on geometrically complex regions (the
    foliage, object silhouettes) and is low on smooth surfaces. On the same view
    and color scale, our model shows lower traversal counts than Radiant Foam
    across most image regions, reducing the mean from $73$ to $49$ cells per ray.
    The speedup therefore comes from a broad reduction in traversal depth, not
    only from a few isolated easy pixels.}
  \label{fig:cost_map}
  \Description{Three-panel figure with a garden rendering, our per-pixel traversal-count heat map, and Radiant Foam's traversal-count heat map on the same color scale.}
\end{figure*}

The per-ray workload model in \cref{eq:ray-work} separates traversal count from the effective cost of each visited cell. \Cref{fig:cells_per_ray} evaluates the first factor. Our method visits $46.1$ cells per ray on average, compared with $66.9$ for Radiant Foam, reducing the mean traversal count by $31\%$. \Cref{fig:cost_map} visualizes the same quantity per pixel, confirming that the reduction holds across the entire image rather than being concentrated in a few regions.

The reduced traversal count is only one part of the measured speedup. We cumulatively ablate the inference optimizations from \cref{sec:inference-optimizations} in \cref{tab:inference_speed_ablation}. The base renderer uses packed half-precision attributes. The subsequent rows add Morton cell ordering, warp-coherent tiling, and low-contribution cell skipping in that order. These choices reduce the cost of a full viewer-style render without changing the evaluation setup. Each timed frame still starts from the requested camera rays and performs the full traversal and compositing pass.

With the representation fixed, the base renderer achieves $230$ FPS. Morton cell ordering raises throughput to $378$ FPS, warp-coherent tiling to $536$ FPS, and low-contribution cell skipping to $623$ FPS. The full stack therefore gives a $2.7\times$ speedup with no measurable change in PSNR or LPIPS.

\begin{table}[H]
  \centering
  \caption{\textbf{Inference-stack ablation.} The representation is fixed. Starting from the base inference renderer, each subsequent row adds one optimization. Averages are over the seven Mip-NeRF~360 scenes.}
  \label{tab:inference_speed_ablation}
  \small
  \setlength{\tabcolsep}{3.5pt}
  \begin{tabular}{lcccc}
    \toprule
    Configuration & PSNR$\uparrow$ & LPIPS$\downarrow$ & FPS$\uparrow$ & Speedup \\
    \midrule
    Base (packed fp16) & 28.98 & 0.235 & 230 & 1.00$\times$ \\
    + Morton cell ordering & 28.98 & 0.235 & 378 & 1.64$\times$ \\
    + warp-coherent tiling & 28.98 & 0.235 & 536 & 2.33$\times$ \\
    + cell skip $(10^{-3})$ & 28.98 & 0.235 & 623 & 2.71$\times$ \\
    \bottomrule
  \end{tabular}
\end{table}

The same renderer raises Radiant Foam throughput from $194$ to $384$ FPS (\cref{tab:cascade}), a smaller gain than for our representation. This difference reflects a shift in the performance bottleneck. As traversal and scheduling overheads decrease, representation evaluation occupies a larger share of rendering time. Our octahedral textures reduce appearance memory traffic, while the learned opacity field concentrates contributions near surfaces and leaves more cells below the skip threshold. The supplementary material analyzes the threshold's speed--quality trade-off.

\subsection{Ablation Studies}
\label{sec:ablations}

To better understand the components introduced in \cref{sec:method}, we ablate the main design choices of our method on the Mip-NeRF~360 scenes. Since these choices are interdependent, we start from the Radiant Foam baseline and add each component in sequence, progressively building up to the final model. This cumulative view shows how each design choice affects both reconstruction quality and rendering cost. \Cref{tab:cascade} reports PSNR and LPIPS for appearance, cells per ray for traversal cost, and FPS on the RTX~5090.

\begin{table}[t]
  \centering
  \caption{\textbf{Cumulative ablation from RadFoam to \methodname{}
    on Mip-NeRF~360.} Each row adds one component to the previous row.
    From the second row onward, all variants use the same deployed inference
    stack, so later speed changes reflect the representation and training
    choices. Points are approximate for Radiant Foam and fixed for our variants.}
  \label{tab:cascade}
  \small
  \setlength{\tabcolsep}{0pt}
  \begin{tabular*}{\columnwidth}{@{\extracolsep{\fill}}lccccc@{}}
    \toprule
    Method & points & PSNR$\uparrow$ & LPIPS$\downarrow$ & cells/ray$\downarrow$ & FPS$\uparrow$ \\
    \midrule
    RadFoam original (SH)     & 4.2/2.1M & 28.44 & 0.277 & 70.2 & 194 \\
    \;\; + our inference stack  & 4.2/2.1M & 28.37 & 0.279 & 70.2 & 384 \\
    \;\; + fixed dense init. & 2M       & 27.90 & 0.290 & 59.6 & 430 \\
    \;\; + exp. density     & 2M       & 28.18 & 0.284 & 51.7 & 496 \\
    \;\; + distortion loss  & 2M       & 28.23 & 0.280 & 52.1 & 475 \\
    \addlinespace[2pt]
    \multicolumn{6}{@{}l}{\emph{Octahedral texture formulation:}} \\
    \;\; + diffuse surface texture & 2M & 27.47 & 0.277 & 56.2 & 463 \\
    \;\; + diffuse mean-pull    & 2M & 27.52 & 0.278 & 52.8 & 552 \\
    \;\; + view-dependent texture  & 2M & 28.18 & 0.250 & 50.8 & 594 \\
    \;\; + vd regularizer = \textbf{\methodname{}}
                                & 2M & \textbf{28.98} & \textbf{0.235} & 50.7 & \textbf{623} \\
    \bottomrule
  \end{tabular*}
\end{table}

Because the goal of our method is to improve rendering speed without sacrificing reconstruction quality, we focus on how the ablated components change both metrics. The first row of \cref{tab:cascade} separates renderer implementation from representation changes. Applying our deployed inference stack, discussed in \cref{sec:speed_analysis}, to the Radiant Foam representation nearly doubles throughput, from $194$ to $384$ FPS, with little change in quality or traversal count.

The dense-initialization ablation replaces the adaptive Radiant Foam training pipeline with a fixed budget of $2$M points. In comparison, Radiant Foam uses scene-dependent point counts of approximately $4.2$M and $2.1$M after densification and pruning. This change intentionally removes discrete adaptive operations from training, including densification, pruning, and variable downsampling, so the following ablations can be studied without relying on these operations. The cost of this simplification is visible in the reconstruction metrics. PSNR drops from $28.37$ to $27.90$ dB before the remaining components are added, which is expected given the substantially smaller and fixed cell budget.

\begin{figure}[t]
  \centering
  \includegraphics[width=\columnwidth]{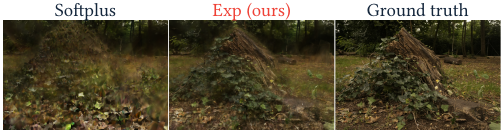}
  \caption{\textbf{Early-training appearance, exp vs.\ softplus.} The same held-out
    view of the \textsc{stump} scene rendered after only $1000$ training steps. With
    softplus the surface is still hazy and washed out; the scale-invariant
    exponential parameterization has already recovered sharp geometry and color
    much closer to the ground truth.}
  \label{fig:activation_render_panel}
  \Description{Three rendered crops of the stump scene after early training: a hazy softplus-density result, a sharper exponential-density result, and the ground-truth image.}
\end{figure}

Replacing the softplus density used by Radiant Foam with our exponential density parameterization reveals an important coupling between quality and speed. The parameterization better matches the optimization of opacity in small cells, allowing the model to converge toward a more surface-like opacity distribution. This raises PSNR from $27.90$ to $28.18$ dB, reduces traversal from $59.6$ to $51.7$ cells per ray, and increases speed from $430$ to $496$ FPS. The faster early convergence is visible in \cref{fig:activation_render_panel}, which compares the softplus and exponential parameterizations after $1000$ of $20000$ training steps.

\paragraph{Surface textures.}
The lower block of \cref{tab:cascade} replaces the spherical-harmonics appearance model with our octahedral texture formulation. As discussed in \cref{sec:considerations}, spherical harmonics require loading many coefficients per cell, while our texture lookups load a smaller fixed set of aligned values. Across the texture block, PSNR increases from $28.23$ to $28.98$ dB, LPIPS decreases from $0.280$ to $0.235$, surpassing Radiant Foam, and FPS increases from $475$ to $623$.

The diffuse surface texture is the component that breaks the direct coupling between spatial detail and cell count. Unlike spherical harmonics, which assign one color field over viewing direction to the whole cell, the diffuse map can vary across the cell footprint. This lets larger cells represent within-cell detail and makes the lower fixed cell budget viable. However, the diffuse texture alone is not sufficient to replace spherical harmonics. This is expected, since the SH baseline already models view-dependent appearance, while a purely diffuse surface texture cannot explain specular highlights or other directional effects. The two texture maps are therefore coupled. The diffuse map provides spatial variation across the cell surface, while the view-dependent map restores the directional residual needed to match the expressiveness of the SH model. Interestingly, the diffuse-only variant is also slower than the SH row, despite requiring fewer appearance loads. Without view-dependent color, the optimizer compensates for missing directional effects by assigning non-negligible opacity to more cells. Fewer ray-cell interactions then fall below the inference skip threshold, offsetting the cheaper diffuse lookup. The mean-pull loss has little effect on visual metrics, but improves throughput from $463$ to $552$ FPS. This suggests that the loss mainly changes the learned opacity distribution, reducing wasted traversal and making the inference-time cell skipping more effective.

The view-dependent texture is the most revealing case. Relative to the diffuse-only texture, it adds a second texture map and therefore increases the appearance data loaded for cells whose color is evaluated. Nevertheless, speed increases from $552$ to $594$ FPS, while LPIPS improves from $0.278$ to $0.250$. The explanation is that the richer appearance model lets the optimizer fit highlights and other view-dependent residuals without spreading opacity over many semi-transparent cells. As a result, the model traverses fewer cells and leaves more low-contribution cells that can be skipped during inference. The view-dependent regularizer then mainly improves reconstruction quality by keeping this component additive and localized, reaching $28.98$ dB PSNR and $0.235$ LPIPS while further increasing speed to $623$ FPS.

\begin{figure}[t]
  \centering
  \includegraphics[width=0.92\columnwidth]{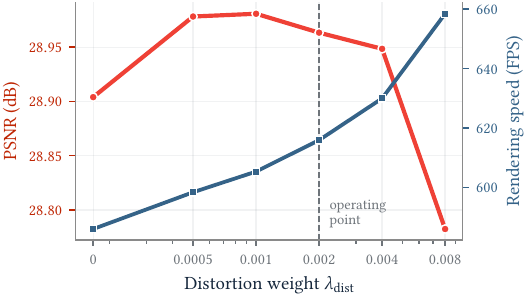}
  \caption{\textbf{Distortion loss trades quality for speed.} As the distortion
    weight $\lambda_{\mathrm{dist}}$ grows, opacity concentrates onto thinner,
    more opaque surfaces (\cref{fig:opacity_hist}), so rays terminate after fewer
    cells and rendering speeds up. We operate at the marked point, which keeps near-peak quality while
    already capturing most of the speedup.}
  \label{fig:distortion_tradeoff}
  \Description{Line plot sweeping the distortion-loss weight, showing PSNR on one axis and rendering speed on the other, with the chosen operating point marked.}
\end{figure}

\paragraph{Distortion weight.}
The distortion loss (\cref{sec:surface-opacity}) acts as a continuous speed and quality control rather than a binary component. \Cref{fig:distortion_tradeoff} sweeps its weight. Larger values concentrate opacity into thinner surfaces, causing rays to terminate sooner and increasing rendering speed. Pushing the weight too far eventually degrades quality because the representation can no longer allocate sufficient thickness to uncertain or semi-transparent regions. The full model operates near the quality peak, while already capturing most of the available speedup. Its effect on the learned opacity distribution is shown in \cref{fig:opacity_hist}.

\paragraph{Texture resolution.}
\Cref{tab:texture_resolution} varies the octahedral diffuse-map resolution independently of the cascade. Quality improves monotonically from $4\times4$ to $8\times8$ texels, most visibly in LPIPS, which decreases from $0.254$ to $0.235$. Higher resolutions come at a quadratic memory cost, trading memory footprint for fidelity. The ablation shows that even the lower-resolution textures retain most of the quality, making $4\times4$ a viable option when memory is constrained. Rendering speed remains essentially unchanged because each evaluated cell still fetches only four texels per texture map for bilinear interpolation. Increasing the map resolution mainly changes storage and, to a lesser extent, cache reuse between nearby rays. We therefore use $8\times8$ maps in the full model.

\begin{table}[t]
  \centering
  \caption{\textbf{Texture resolution ablation on Mip-NeRF~360.}
    Mean reconstruction quality over the 7 scenes and rendering speed on an
    RTX~5090. Increasing the octahedral diffuse-map resolution mainly improves
    LPIPS, while rendering speed remains nearly unchanged.}
  \label{tab:texture_resolution}
  \small
  \setlength{\tabcolsep}{5pt}
  \begin{tabular}{@{}l cccc@{}}
    \toprule
    Resolution & PSNR$\uparrow$ & SSIM$\uparrow$ & LPIPS$\downarrow$ & FPS$\uparrow$ \\
    \midrule
    $4\times4$ & 28.85 & 0.842 & 0.254 & 619 \\
    $6\times6$ & 28.94 & 0.846 & 0.243 & 615 \\
    $8\times8$ (ours) & \textbf{28.98} & \textbf{0.848} & \textbf{0.235} & 623 \\
    \bottomrule
  \end{tabular}
\end{table}

\subsection{Applications of Ray-Based Rendering}
\label{sec:applications}

The quantitative results above use standard pinhole benchmark views so that quality and speed are directly comparable to prior work. The practical motivation for retaining ray-based image formation is broader because camera effects can be expressed by changing the rays supplied to the renderer rather than by modifying a projection, sorting, or splatting pipeline. For a fisheye or lens-distorted camera, each pixel simply emits the ray prescribed by the lens model. For depth of field, a pixel emits a small bundle of rays sampled over an aperture and averages their returned colors. The traversal, appearance lookup, and compositing code are unchanged.

\begin{figure}[t]
  \centering
  \includegraphics[width=\columnwidth,keepaspectratio]{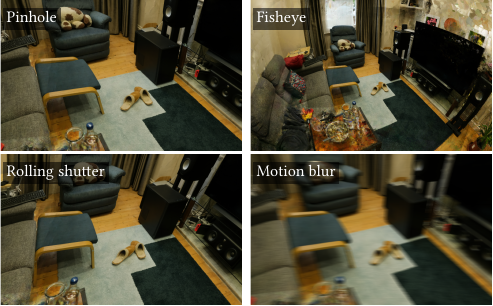}
  \caption{\textbf{Ray-based camera effects.} All panels render the same trained
    representation while changing only the rays sent to the renderer. The examples
    show a standard pinhole camera, a wide-angle fisheye camera, row-dependent poses for
    rolling shutter, and temporal averaging for camera motion blur.}
  \label{fig:ray_applications}
  \Description{Four-panel figure showing the same scene rendered with pinhole, fisheye, rolling-shutter, and motion-blur camera effects.}
\end{figure}

The same interface also supports effects that vary the camera over the image or over time. For rolling shutter, rays are generated from row-dependent camera poses. For camera motion blur, several time samples are rendered and averaged. Fisheye and rolling-shutter rendering retain one ray per pixel, whereas depth of field and motion blur submit multiple rays per pixel and therefore increase the total frame workload. The reported pinhole throughput therefore does not apply directly to these multisampled effects. \Cref{fig:ray_applications} shows pinhole, fisheye, rolling-shutter, and camera-motion-blur renders from a single trained representation, while \cref{fig:cover} shows a depth-of-field-style aperture rendering. These examples are intended to demonstrate renderer flexibility; the controlled quantitative comparison remains the Mip-NeRF~360 benchmark above.

To evaluate performance beyond desktop GPUs, we deploy the renderer on an iPhone~16. The same Garden representation used in the RTX~5090 evaluation renders at an average of $40$ FPS at a resolution with $1024$ pixels along the longer image dimension. This deployment uses the original representation without compression, cell-count reduction, or other mobile-specific simplifications and accounts for the full rendering pipeline and display output.

\section{Discussion and Conclusions}

We introduced a ray-based reconstruction method designed for efficient GPU execution. Our results show that a ray-based pipeline designed around inference cost and GPU execution can exceed the rendering throughput of popular rasterized methods while retaining the flexibility of ray-based image formation. The same renderer supports fisheye projection, rolling shutter, depth of field, and motion blur through changes to ray generation and sampling. We additionally demonstrate interactive rendering on a mobile device (\cref{sec:applications}).

Building on Radiant Foam, our method traverses a Voronoi diagram through local cell adjacency rather than repeated global BVH intersection queries. We analyze the resulting rendering cost in terms of cells visited per ray, work performed in each cell, and memory coherence across neighboring rays. To improve GPU throughput, we reorder cells in Morton order, schedule rays in compact screen-space tiles, and skip appearance evaluation for cells whose maximum possible contribution falls below a fixed threshold. At the representation level, this analysis motivates reducing both the number of cells visited and the cost of appearance evaluation in each cell.

To reduce traversal, we add the Mip-NeRF~360 distortion loss. This regularizer concentrates rendering weights near surfaces and promotes a bimodal opacity distribution. Low-contribution appearance evaluations can then be skipped in transparent free space, while opaque surfaces terminate rays early. To reduce per-cell appearance cost, we replace spherical harmonics with paired octahedral surface and view-dependent textures. These halve the appearance values required for each cell evaluation relative to spherical harmonics, while the surface texture allows color to vary across a cell's image footprint. This within-cell variation lets larger cells represent spatial detail that would otherwise require additional cells.

We also introduce a scale-invariant density parameterization that gives cells of equal opacity identically scaled gradients regardless of size. Together with dense feature-matching initialization, this supports a fixed-budget training pipeline. We optimize site positions, densities, and textures at the target image resolution for the full training run, without pruning, densification, or progressive-resolution schedules.

\paragraph{Limitations and future work.}
The fixed cell budget remains a limitation in regions observed by few training views, where dense feature matching may provide too few reliable sites. Although our method often preserves more background detail than Radiant Foam and PowerFoam, an adaptive cell-allocation strategy could place additional capacity in these underrepresented regions. A second direction is explicit surface extraction. The Voronoi diagram, surface-concentrated opacity, and surface-indexed textures may provide a useful starting point for extracting a textured mesh compatible with standard rendering pipelines.

Taken together, our results show that an explicit Voronoi radiance field can exceed the rendering throughput of rasterized alternatives while retaining comparable reconstruction quality and the flexibility of ray-based image formation. More broadly, these findings suggest that real-time ray-based rendering depends on co-designing traversal, representation, optimization, and GPU execution rather than on the choice between ray tracing and rasterization alone.

\begin{acks}

This work was partially supported by the Wallenberg AI, Autonomous Systems and Software Program (WASP) funded by the Knut and Alice Wallenberg Foundation. Computational resources were provided by NAISS at \href{https://www.nsc.liu.se/}{NSC Berzelius}, partially funded by the Swedish Research Council, grant agreement no. 2022-06725.

\end{acks}

\bibliographystyle{ACM-Reference-Format}
\bibliography{references}

\clearpage
\setcounter{page}{1}
\appendix
\section{Implementation and Training Details}
\label{supp:implementation}

All models reported in the paper are trained on a single RTX~5090 GPU for $20{,}000$ iterations with a fixed budget of $2$M Voronoi sites. Each iteration samples $1$M rays uniformly over the training images and pixels. We use the standard Mip-NeRF~360 downsampling factors, training outdoor scenes at one-quarter resolution and indoor scenes at one-half resolution. These resolutions remain fixed throughout training. Both the view-independent and view-dependent octahedral textures have a resolution of $8\times8$ texels per cell. We use a white background and do not use patch sampling, an SSIM loss, pruning, or densification.

\begin{table}[h]
  \centering
  \caption{Training settings used for the Mip-NeRF~360 experiments. Learning-rate entries give the schedule endpoints.}
  \label{tab:supp-training-settings}
  \begin{tabular}{lcc}
    \toprule
    Setting & Outdoor & Indoor \\
    \midrule
    Image downsampling & $4\times$ & $2\times$ \\
    Site-position learning rate & $2\mathord{\times}10^{-4} \rightarrow 5\mathord{\times}10^{-6}$ & same \\
    Density learning rate & $10^{-1} \rightarrow 10^{-2}$ & same \\
    Texture learning rate & $2\mathord{\times}10^{-2} \rightarrow 5\mathord{\times}10^{-4}$ & same \\
    View-dependent ramp length & $4{,}000$ & $0$ \\
    $\lambda_{\mathrm{dist}}$ & $2\mathord{\times}10^{-3}$ & same \\
    $\lambda_{\mathrm{vd}}$ & $10^{-2}$ & $10^{-4}$ \\
    $\lambda_{\mathrm{mean}}$ & $5\mathord{\times}10^{-3}$ & $10^{-4}$ \\
    \bottomrule
  \end{tabular}
\end{table}

We use Adam with $\beta_1=0.9$, $\beta_2=0.999$, and $\epsilon=10^{-15}$. Site positions follow a cosine learning-rate schedule for the first $18{,}000$ iterations and are then fixed. Density uses a linear warmup for $2{,}000$ iterations followed by cosine decay, while both texture learning rates use cosine decay throughout training. For outdoor scenes, the view-dependent texture learning rate is additionally multiplied by a quadratic ramp that reaches one at iteration $4{,}000$. View-dependent appearance is optimized from the first iteration for indoor scenes.

The $20{,}000$ optimization iterations take $33.2$--$50.1$ minutes per scene and $39.9$ minutes on average over the seven Mip-NeRF~360 scenes. The one-time RoMa v2 correspondence-based point-cloud construction takes $41.7$ seconds per scene on average.

The density-aware initialization described in the main paper uses a $128^3$ occupancy grid and inverse occupancy as the sampling probability. We add $5{,}000$ random background sites to the sampled point cloud and perturb sampled site positions with zero-mean Gaussian noise of standard deviation $10^{-2}$. Because the sites move during optimization, we periodically update the Delaunay triangulation and its Voronoi adjacency.

\section{Benchmarking Protocol}
\label{supp:benchmarking}

All timings for our method are measured on a single RTX~5090 GPU. We time full-frame rendering after model loading and one-time, view-independent preprocessing. The timed region follows the use case of an interactive viewer. A camera pose is requested, rays are generated for the full image, and the renderer performs Voronoi traversal, appearance evaluation, alpha compositing, and output generation for that frame.
We do not use precomputations that require knowledge of future renders or the future camera path.

For each held-out view, we first render one image for the quality metrics. Runtime is then measured using CUDA events after warmup iterations, and the per-view time is averaged over repeated full-frame renders. The benchmark performs 10 warmup iterations and 20 timed iterations per view. Frames per second are computed as the reciprocal of the mean milliseconds per frame over the held-out views.

\subsection{Metric Computation}
\label{supp:metrics}

We report PSNR, SSIM, and LPIPS on the held-out Mip-NeRF~360 views. Predictions and references are collected as lossless RGB PNGs and evaluated afterward with one implementation shared by all methods. PSNR and SSIM operate on RGB values in $[0,1]$, and LPIPS uses the VGG backbone with normalized inputs.

\section{Baseline Details}
\label{supp:baselines}

We reproduce the baselines in \cref{tab:mip360_raytraced,tab:mip360_rasterized} using their publicly released implementations. We retain each method's published training configuration without method-specific retuning. Quality evaluation uses the standard Mip-NeRF~360 resolutions, with the provided \texttt{images\_4} JPEGs for outdoor scenes and \texttt{images\_2} JPEGs for indoor scenes. Held-out views follow the \texttt{llffhold=8} convention and contain every eighth image after sorting the COLMAP image names.

Baseline timings use the same RTX~5090 GPU as our method. Methods using our common CUDA helper receive $10$ warmup renders followed by $10$ complete passes over the held-out cameras, measured with CUDA events. These measurements model an interactive viewer in which each frame may be requested from an arbitrary camera pose. We therefore include per-view camera setup and any query required to begin rendering that pose. Model loading, static acceleration-structure construction, reference-image access, metric evaluation, and image encoding are excluded.

\subsection{Method-Specific Evaluation Details}

\paragraph{Instant-NGP}
We evaluate the base and big configurations used in the 3DGS comparison with the latest revision of the official Instant-NGP implementation available~\cite{muller2022instant,kerbl20233d}. Instant-NGP predates Mip-NeRF~360, and this evaluation therefore uses dataset support added to the repository after the original publication. Our reproduced PSNR is approximately $0.7$~dB higher for the base configuration and $1.2$~dB higher for the big configuration than the values quoted in the 3DGS supplement. We report our reproduced values under the shared split and metric implementation.

\paragraph{Zip-NeRF}
Zip-NeRF~\cite{barron2023zip} is included as a high-quality NeRF reference rather than as a single-GPU training baseline. We train it with the official Zip-NeRF implementation on a $4\times$ GH200 setup, because the published configuration is substantially more expensive than the methods otherwise evaluated on the RTX~5090. We evaluate the validation renders from this run with the same metric implementation and held-out split used for the other methods. To obtain an indicative rendering speed, we run the official Zip-NeRF renderer on the RTX~5090 with an updated JAX runtime required for compatibility with this GPU. The reported FPS is therefore an approximate local inference measurement and should not be interpreted as a matched single-GPU training-and-rendering reproduction.

\paragraph{3DGRT and 3DGUT}
We use the official 3DGRUT implementation and the authors' Mip-NeRF~360 configurations for both methods~\cite{moenne20243d,wu20253dgut}. Whereas the upstream benchmark measures the central rendering kernel, our timed path includes per-view camera and sensor construction, rendering, RGBA extraction, and background compositing. View-independent activations of the frozen Gaussian parameters are cached outside the timed region.

\paragraph{Radiant Foam}
We use the official implementation with the authors' Mip-NeRF~360 configurations~\cite{govindarajan2025radiant}. The upstream benchmark determines the starting cells for all validation cameras before timing, whereas we include the nearest-cell query for each view in the timed region, as that reflects the actual usage scenario of an interactive viewer.

\paragraph{PowerFoam}
We use the official implementation with the published indoor and outdoor Mip-NeRF~360 configurations~\cite{govindarajan2026power}. We report its rasterized and ray-traced renderers separately. The two renderers match in reconstruction quality, so their quality metrics are identical, while their rendering speeds are measured separately. For ray tracing, the timed region includes the nearest weighted-cell query for each requested camera.

\paragraph{3D Gaussian Splatting}
We use the latest revision of the official implementation available when our experiments were run, which incorporates fixes and improvements made after the original release~\cite{kerbl20233d}. We retain the published Mip-NeRF~360 recipe and default black background. Since the repository does not provide an end-to-end rendering benchmark, we time its standard render function. Each timed invocation includes the complete per-camera rasterization call.

\paragraph{Radiance Meshes}
The authors release Python/CUDA code for training and evaluation and a separate VKRM Vulkan renderer used for the throughput measurements in their paper~\cite{mai2026radiance}. We train with the Python/CUDA implementation and evaluate both its standard render function and the Vulkan renderer using the exported scene. In our experiments, the Vulkan renderer is substantially faster but produces substantially lower image quality than the Python/CUDA renderer. We consulted the authors and attempted to reconcile the two outputs but did not identify a configuration that reproduced the Python/CUDA quality with the Vulkan renderer. We therefore report each backend as a separate row and do not combine the quality of the Python/CUDA renderer with the throughput of the Vulkan renderer.

\paragraph{Triangle Splatting}
We use the official implementation with the authors' published Mip-NeRF~360 settings~\cite{held2026triangle}. 

\section{Additional Application Results}
\label{supp:applications}

The application examples use the same representations trained for the pinhole benchmark. Only ray generation changes. Fisheye and lens distortion alter pixel ray directions, rolling shutter changes the camera pose by image row, depth of field samples ray origins over an aperture, and motion blur samples camera poses over the exposure interval. The Voronoi traversal, appearance evaluation, and compositing are unchanged.

\begin{figure}[h]
  \centering
  \includegraphics[width=0.88\columnwidth]{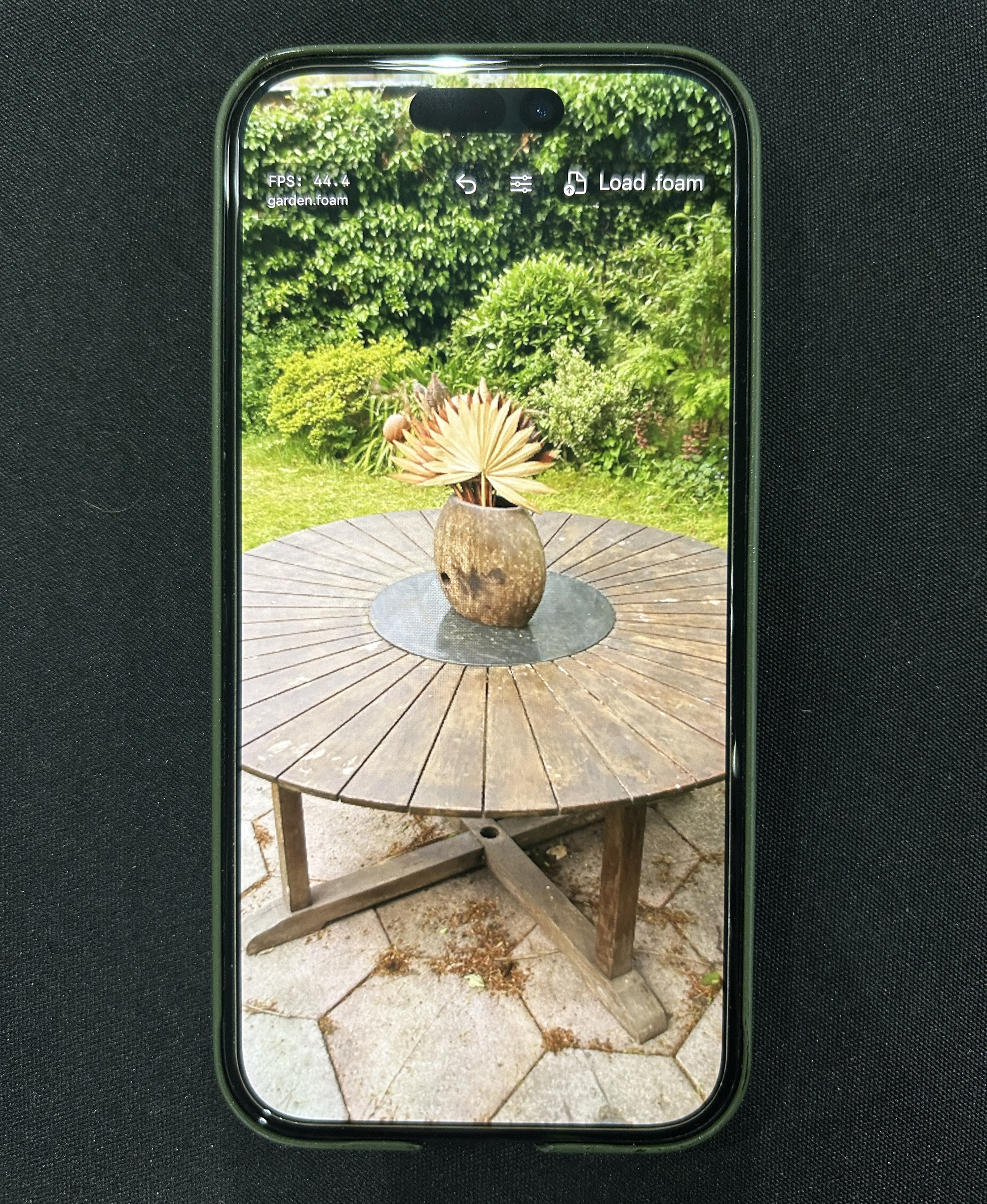}
  \caption{\textbf{Mobile viewer demonstration.} The Garden representation runs interactively on an iPhone~16. It uses the same Garden model with $2$M sites as the desktop evaluation, without compression, a lower cell count, or retraining.}
  \label{fig:supp-mobile-demo}
  \Description{Photograph of an iPhone running an interactive viewer that displays the Garden scene, with an on-screen frame-rate counter.}
\end{figure}

For the mobile timing reported in the main paper, we port the renderer to equivalent Metal compute kernels and embed it in a viewer application that supports interactive camera motion. We render the Garden scene with $1024$ pixels along the longer image dimension and include the complete rendering pipeline and display output.

\section{Per-Scene Results}
\label{supp:per-scene-results}

Per-scene PSNR, SSIM, LPIPS, and rendering speed for the Mip-NeRF~360 scenes are reported in \cref{tab:supp-per-scene-psnr,tab:supp-per-scene-ssim,tab:supp-per-scene-lpips,tab:supp-per-scene-fps}.

\begin{table*}[p]
  \centering
  \small
  \setlength{\tabcolsep}{5.2pt}
  \renewcommand{\arraystretch}{0.95}
  \caption{\textbf{Per-scene PSNR$\uparrow$ on Mip-NeRF~360.}
  Per column, \colorbox{tbf}{best}, \colorbox{tbs}{second}, and \colorbox{tbt}{third}. }
  \label{tab:supp-per-scene-psnr}
  \begin{tabular}{@{}l!{\vrule width 0.35pt}rrrr!{\vrule width 0.35pt}rrr!{\vrule width 0.35pt}r@{}}
    \toprule
    & \multicolumn{4}{c!{\vrule width 0.35pt}}{Indoor} & \multicolumn{3}{c!{\vrule width 0.35pt}}{Outdoor} & \\
    \cmidrule(lr){2-5}\cmidrule(lr){6-8}
    Method & Room & Counter & Bonsai & Kitchen & Bicycle & Garden & Stump & Mean \\
    \midrule
    Zip-NeRF & \cellcolor{tbf}32.97 & 29.06 & \cellcolor{tbf}34.73 & \cellcolor{tbf}32.42 & \cellcolor{tbf}25.87 & \cellcolor{tbf}28.19 & \cellcolor{tbf}27.32 & \cellcolor{tbf}30.08 \\
    iNGP (base) & 30.37 & 27.85 & 31.27 & 24.85 & 24.00 & 25.58 & 25.89 & 27.12 \\
    iNGP (big) & \cellcolor{tbs}31.97 & 28.54 & 32.14 & 25.17 & 24.52 & 26.73 & \cellcolor{tbs}26.78 & 27.98 \\
    3DGRT & 30.08 & 28.37 & 31.59 & 29.68 & 24.73 & 26.73 & 26.23 & 28.20 \\
    RadFoam & 30.82 & 28.56 & 32.26 & 31.23 & 24.21 & 26.54 & 25.45 & 28.44 \\
    PowerFoam & 31.07 & \cellcolor{tbf}29.66 & \cellcolor{tbt}33.20 & 31.32 & 24.10 & 26.92 & 25.17 & 28.78 \\
    3DGS & 31.59 & \cellcolor{tbt}29.10 & 32.40 & 31.36 & \cellcolor{tbs}25.21 & \cellcolor{tbs}27.48 & \cellcolor{tbt}26.66 & \cellcolor{tbs}29.11 \\
    3DGUT & \cellcolor{tbt}31.68 & 29.08 & 32.48 & 31.03 & \cellcolor{tbt}25.05 & 27.16 & 26.41 & \cellcolor{tbt}28.98 \\
    Radiance Meshes & 30.51 & 28.51 & 30.83 & 30.80 & 24.98 & 26.98 & 26.37 & 28.43 \\
    Radiance Meshes (Vulkan) & 28.35 & 27.10 & 28.57 & 28.28 & 22.37 & 24.74 & 24.84 & 26.32 \\
    Triangle Splatting & 31.12 & 28.84 & 32.00 & \cellcolor{tbt}31.37 & 24.77 & \cellcolor{tbt}27.20 & 26.13 & 28.78 \\
    \textbf{\methodname{}} & 31.23 & \cellcolor{tbs}29.48 & \cellcolor{tbs}33.42 & \cellcolor{tbs}31.54 & 24.29 & 26.98 & 25.94 & \cellcolor{tbt}28.98 \\
    \bottomrule
  \end{tabular}
\end{table*}

\begin{table*}[p]
  \centering
  \small
  \setlength{\tabcolsep}{5.2pt}
  \renewcommand{\arraystretch}{0.95}
  \caption{\textbf{Per-scene SSIM$\uparrow$ on Mip-NeRF~360.}
  Per column, \colorbox{tbf}{best}, \colorbox{tbs}{second}, and \colorbox{tbt}{third}. }
  \label{tab:supp-per-scene-ssim}
  \begin{tabular}{@{}l!{\vrule width 0.35pt}rrrr!{\vrule width 0.35pt}rrr!{\vrule width 0.35pt}r@{}}
    \toprule
    & \multicolumn{4}{c!{\vrule width 0.35pt}}{Indoor} & \multicolumn{3}{c!{\vrule width 0.35pt}}{Outdoor} & \\
    \cmidrule(lr){2-5}\cmidrule(lr){6-8}
    Method & Room & Counter & Bonsai & Kitchen & Bicycle & Garden & Stump & Mean \\
    \midrule
    Zip-NeRF & \cellcolor{tbf}0.929 & \cellcolor{tbt}0.906 & \cellcolor{tbf}0.952 & \cellcolor{tbf}0.930 & \cellcolor{tbf}0.773 & \cellcolor{tbs}0.863 & \cellcolor{tbf}0.788 & \cellcolor{tbf}0.877 \\
    iNGP (base) & 0.869 & 0.829 & 0.902 & 0.762 & 0.564 & 0.690 & 0.654 & 0.753 \\
    iNGP (big) & 0.894 & 0.853 & 0.921 & 0.785 & 0.617 & 0.765 & 0.709 & 0.792 \\
    3DGRT & 0.903 & 0.901 & 0.937 & 0.914 & 0.744 & 0.847 & 0.765 & 0.859 \\
    RadFoam & 0.901 & 0.875 & 0.925 & 0.906 & 0.674 & 0.810 & 0.709 & 0.829 \\
    PowerFoam & 0.907 & 0.895 & 0.937 & 0.915 & 0.675 & 0.824 & 0.690 & 0.835 \\
    3DGS & \cellcolor{tbs}0.920 & \cellcolor{tbs}0.909 & 0.942 & \cellcolor{tbs}0.927 & \cellcolor{tbs}0.765 & \cellcolor{tbf}0.868 & \cellcolor{tbt}0.771 & \cellcolor{tbs}0.872 \\
    3DGUT & 0.918 & \cellcolor{tbf}0.910 & \cellcolor{tbt}0.944 & \cellcolor{tbs}0.927 & \cellcolor{tbt}0.760 & 0.856 & \cellcolor{tbt}0.771 & \cellcolor{tbt}0.870 \\
    Radiance Meshes & 0.913 & 0.905 & 0.941 & 0.916 & 0.744 & 0.849 & 0.755 & 0.861 \\
    Radiance Meshes (Vulkan) & 0.833 & 0.823 & 0.883 & 0.850 & 0.598 & 0.724 & 0.624 & 0.762 \\
    Triangle Splatting & \cellcolor{tbt}0.919 & \cellcolor{tbf}0.910 & \cellcolor{tbs}0.946 & \cellcolor{tbt}0.926 & 0.749 & \cellcolor{tbt}0.858 & \cellcolor{tbs}0.772 & 0.869 \\
    \textbf{\methodname{}} & 0.916 & 0.892 & 0.939 & 0.918 & 0.711 & 0.826 & 0.733 & 0.848 \\
    \bottomrule
  \end{tabular}
\end{table*}

\begin{table*}[p]
  \centering
  \small
  \setlength{\tabcolsep}{5.2pt}
  \renewcommand{\arraystretch}{0.95}
  \caption{\textbf{Per-scene LPIPS$\downarrow$ on Mip-NeRF~360.}
  Per column, \colorbox{tbf}{best}, \colorbox{tbs}{second}, and \colorbox{tbt}{third}. Lower is better.}
  \label{tab:supp-per-scene-lpips}
  \begin{tabular}{@{}l!{\vrule width 0.35pt}rrrr!{\vrule width 0.35pt}rrr!{\vrule width 0.35pt}r@{}}
    \toprule
    & \multicolumn{4}{c!{\vrule width 0.35pt}}{Indoor} & \multicolumn{3}{c!{\vrule width 0.35pt}}{Outdoor} & \\
    \cmidrule(lr){2-5}\cmidrule(lr){6-8}
    Method & Room & Counter & Bonsai & Kitchen & Bicycle & Garden & Stump & Mean \\
    \midrule
    Zip-NeRF & \cellcolor{tbf}0.238 & \cellcolor{tbs}0.223 & \cellcolor{tbf}0.195 & \cellcolor{tbs}0.133 & \cellcolor{tbs}0.227 & \cellcolor{tbt}0.127 & \cellcolor{tbf}0.237 & \cellcolor{tbf}0.197 \\
    iNGP (base) & 0.375 & 0.363 & 0.297 & 0.326 & 0.464 & 0.320 & 0.407 & 0.365 \\
    iNGP (big) & 0.319 & 0.321 & 0.259 & 0.290 & 0.411 & 0.256 & 0.360 & 0.316 \\
    3DGRT & 0.298 & 0.256 & 0.241 & 0.166 & 0.256 & 0.145 & \cellcolor{tbt}0.256 & 0.231 \\
    RadFoam & 0.300 & 0.286 & 0.268 & 0.192 & 0.370 & 0.190 & 0.334 & 0.277 \\
    PowerFoam & 0.295 & 0.256 & 0.244 & 0.170 & 0.340 & 0.161 & 0.334 & 0.257 \\
    3DGS & 0.284 & 0.256 & 0.253 & \cellcolor{tbt}0.154 & 0.238 & \cellcolor{tbs}0.122 & \cellcolor{tbs}0.251 & \cellcolor{tbt}0.223 \\
    3DGUT & 0.289 & 0.252 & 0.243 & 0.156 & \cellcolor{tbt}0.235 & 0.143 & \cellcolor{tbt}0.256 & 0.225 \\
    Radiance Meshes & 0.275 & 0.248 & 0.255 & 0.181 & 0.287 & 0.176 & 0.283 & 0.244 \\
    Radiance Meshes (Vulkan) & 0.347 & 0.269 & 0.279 & 0.202 & 0.389 & 0.232 & 0.334 & 0.293 \\
    Triangle Splatting & \cellcolor{tbs}0.247 & \cellcolor{tbf}0.222 & \cellcolor{tbt}0.229 & \cellcolor{tbf}0.124 & \cellcolor{tbf}0.224 & \cellcolor{tbf}0.113 & \cellcolor{tbs}0.251 & \cellcolor{tbs}0.201 \\
    \textbf{\methodname{}} & \cellcolor{tbt}0.264 & \cellcolor{tbt}0.243 & \cellcolor{tbs}0.222 & 0.160 & 0.298 & 0.166 & 0.294 & 0.235 \\
    \bottomrule
  \end{tabular}
\end{table*}

\begin{table*}[p]
  \centering
  \small
  \setlength{\tabcolsep}{5.2pt}
  \renewcommand{\arraystretch}{0.95}
  \caption{\textbf{Per-scene FPS$\uparrow$ on Mip-NeRF~360.}
  Per column, \colorbox{tbf}{best}, \colorbox{tbs}{second}, and \colorbox{tbt}{third}. Frames per second are measured on an RTX~5090.}
  \label{tab:supp-per-scene-fps}
  \begin{tabular}{@{}l!{\vrule width 0.35pt}rrrr!{\vrule width 0.35pt}rrr!{\vrule width 0.35pt}r@{}}
    \toprule
    & \multicolumn{4}{c!{\vrule width 0.35pt}}{Indoor} & \multicolumn{3}{c!{\vrule width 0.35pt}}{Outdoor} & \\
    \cmidrule(lr){2-5}\cmidrule(lr){6-8}
    Method & Room & Counter & Bonsai & Kitchen & Bicycle & Garden & Stump & Mean \\
    \midrule
    Zip-NeRF & 0.19 & 0.19 & 0.19 & 0.19 & 0.30 & 0.28 & 0.29 & 0.23 \\
    iNGP (base) & 25 & 18 & 22 & 32 & 30 & 38 & 21 & 27 \\
    iNGP (big) & 26 & 17 & 22 & 32 & 29 & 37 & 19 & 26 \\
    3DGRT & 111 & 75 & 77 & 45 & 96 & 103 & 112 & 88 \\
    RadFoam & 230 & 221 & 228 & 210 & 51 & 233 & \cellcolor{tbt}185 & 194 \\
    PowerFoam (RT) & 154 & 127 & 137 & 129 & 113 & 148 & 111 & 131 \\
    PowerFoam (rast.) & \cellcolor{tbs}440 & 290 & \cellcolor{tbs}375 & \cellcolor{tbt}295 & \cellcolor{tbt}231 & \cellcolor{tbt}316 & 120 & \cellcolor{tbt}295 \\
    3DGS & 253 & 268 & \cellcolor{tbs}375 & 219 & 112 & 147 & 168 & 220 \\
    3DGUT & 335 & \cellcolor{tbt}316 & 333 & 246 & 162 & 204 & 184 & 254 \\
    Radiance Meshes & 54 & 51 & 49 & 45 & 53 & 56 & 66 & 53 \\
    Radiance Meshes (Vulkan) & \cellcolor{tbt}383 & \cellcolor{tbs}335 & \cellcolor{tbt}337 & \cellcolor{tbs}325 & \cellcolor{tbs}360 & \cellcolor{tbs}393 & \cellcolor{tbs}200 & \cellcolor{tbs}333 \\
    Triangle Splatting & 160 & 181 & 233 & 141 & 112 & 123 & 109 & 151 \\
    \textbf{\methodname{}} & \cellcolor{tbf}585 & \cellcolor{tbf}589 & \cellcolor{tbf}559 & \cellcolor{tbf}595 & \cellcolor{tbf}558 & \cellcolor{tbf}825 & \cellcolor{tbf}649 & \cellcolor{tbf}623 \\
    \bottomrule
  \end{tabular}
\end{table*}

\section{Additional Ablations}
\label{supp:ablations}

\begin{table}[H]
  \centering
  \caption{\textbf{Cell-skip threshold sweep.} Morton ordering and warp-coherent tiling are enabled for all rows. Increasing the threshold skips more low-contribution texture evaluations, trading quality for speed.}
  \label{tab:cell_skip_sweep}
  \small
  \setlength{\tabcolsep}{4pt}
  \begin{tabular}{lcccc}
    \toprule
    Threshold & PSNR$\uparrow$ & $\Delta$PSNR & LPIPS$\downarrow$ & FPS$\uparrow$ \\
    \midrule
    0 & 28.98 & +0.00 & 0.235 & 538 \\
    $10^{-5}$ & 28.98 & +0.00 & 0.235 & 597 \\
    $10^{-4}$ & 28.98 & +0.00 & 0.235 & 609 \\
    $3\times 10^{-4}$ & 28.98 & +0.00 & 0.235 & 613 \\
    $10^{-3}$ & 28.98 & -0.00 & 0.235 & 623 \\
    $3\times 10^{-3}$ & 28.93 & -0.04 & 0.234 & 631 \\
    $10^{-2}$ & 28.17 & -0.79 & 0.240 & 649 \\
    \bottomrule
  \end{tabular}
\end{table}

\Cref{tab:cell_skip_sweep} reports the inference-time threshold sweep used for the headline renderer. The selected threshold of $10^{-3}$ preserves PSNR and LPIPS at the reported precision while increasing throughput from $538$ to $623$ FPS relative to evaluating appearance in every traversed cell.

\end{document}